\documentclass[10pt,journal,compsoc]{IEEEtran}
\ifCLASSOPTIONcompsoc
  \usepackage[nocompress]{cite}
\else
  \usepackage{cite}
\fi
\ifCLASSINFOpdf
   \usepackage[pdftex]{graphicx}
   \usepackage{subcaption}
\else
\fi
\usepackage{pdfpages}
\usepackage{amsmath,amsfonts,bm}
\newcommand{\etal}{{\em et al.}}	
\newcommand{\eg}{{\em e.g.}}		
\newcommand{\ie}{{\em i.e.}}		
\newcommand{\etc}{{\em etc}}		
\newcommand{\wrt}{{\em w.r.t.}}		
\def\gC{{\mathcal{C}}}
\def\gD{{\mathcal{D}}}
\def\gE{{\mathcal{E}}}
\def\gF{{\mathcal{F}}}

\def\gL{{\mathcal{L}}}
\def\gN{{\mathcal{N}}}
\def\gO{{\mathcal{O}}}
\def\gT{{\mathcal{T}}}
\def\gX{{\mathcal{X}}}
\def\rvc{{\mathbf{c}}}
\def\rvh{{\mathbf{h}}}
\def\rvp{{\mathbf{p}}}

\def\rvy{{\mathbf{y}}}
\def\rvz{{\mathbf{z}}}
\def\rmI{{\mathbf{I}}}
\def\rmP{{\mathbf{P}}}
\def\rmX{{\mathbf{X}}}
\def\vzero{{\bm{0}}}
\def\vmu{{\bm{\mu}}}
\newcommand{\vDelta}{{\boldsymbol{\Delta}}}
\newcommand{\vomega}{{\bm{\omega}}}
\newcommand{\vrho}{{\boldsymbol{\rho}}}
\newcommand{\vsigma}{{\bm{\sigma}}}
\newcommand{\vepsilon}{{\bm{\epsilon}}}
\newcommand{\withdim}[1]{{\ \in \mathbb{R}^{#1}}}
\newcommand{\distance}[1]{{\lVert #1 \rVert}}

\DeclareMathOperator*{\argmin}{arg\,min}

\usepackage{booktabs}
\usepackage{multirow}
\usepackage{hyperref}

\begin{document}
%
\title{XADLiME: eXplainable Alzheimer's Disease Likelihood Map Estimation via Clinically-guided Prototype Learning}
%
%
%
%

\author{Ahmad~Wisnu~Mulyadi,
        Wonsik~Jung,
        Kwanseok~Oh,
        Jee~Seok~Yoon,
        and~Heung-Il~Suk,~\IEEEmembership{Senior Member,~IEEE}
        
\IEEEcompsocitemizethanks{
\IEEEcompsocthanksitem A. W. Mulyadi, W. Jung, and J. S. Yoon are with the Department of Brain and Cognitive Engineering, Korea University, Seoul 02841, Republic of Korea (e-mail: \{wisnumulyadi, ssikjeong1, wltjr1007\}@korea.ac.kr).
\IEEEcompsocthanksitem K. Oh is with the Department of Artificial Intelligence, Korea University, Seoul 02841, Republic of Korea (e-mail: ksohh@korea.ac.kr).
\IEEEcompsocthanksitem H.-I. Suk is with the Department of Artificial Intelligence and the Department of Brain and Cognitive Engineering, Korea University, Seoul 02841, Republic of Korea and the corresponding author (e-mail: hisuk@korea.ac.kr).
}}

%
%

\markboth{PREPRINT}%
{Mulyadi \MakeLowercase{\textit{et al.}}: XADLiME: eXplainable Alzheimer's Disease Likelihood Map Estimation via Clinically-guided Prototype Learning}
%



\IEEEtitleabstractindextext{%
\begin{abstract}
Diagnosing Alzheimer's disease (AD) involves a deliberate diagnostic process owing to its innate traits of irreversibility with subtle and gradual progression. These characteristics make AD biomarker identification from structural brain imaging \mbox{(\eg, structural MRI)} scans quite challenging. Furthermore, there is a high possibility of getting entangled with normal aging. We propose a novel deep-learning approach through eXplainable AD Likelihood Map Estimation (XADLiME) for AD progression modeling over 3D sMRIs using clinically-guided prototype learning. Specifically, we establish a set of topologically-aware prototypes onto the clusters of latent clinical features, uncovering an AD spectrum manifold. We then measure the similarities between latent clinical features and well-established prototypes, estimating a ``pseudo" likelihood map. By considering this pseudo map as an enriched reference, we employ an estimating network to estimate the AD likelihood map over a 3D sMRI scan. Additionally, we promote the explainability of such a likelihood map by revealing a comprehensible overview from two perspectives: clinical and morphological. During the inference, this estimated likelihood map served as a substitute over unseen sMRI scans for effectively conducting the downstream task while providing thorough explainable states.
\end{abstract}

}

\maketitle

\IEEEdisplaynontitleabstractindextext

%
\IEEEpeerreviewmaketitle

\IEEEraisesectionheading{\section{Introduction}\label{sec:introduction}}

%
%
%
%
\IEEEPARstart{A}{lzheimer's} disease (AD) is widely acknowledged as a neurodegenerative disease distinguished by being an irreversible, gradually yet subtle progressive, and prevailing cause of dementia. The precursor symptoms commonly manifest during mild cognitive impairment (MCI) as an intermediate stage between cognitive normal (CN) towards AD, which involves declining cognitive functions and recurring memory loss \cite{apostolova2008}. We refer to such deteriorated stage advancement as the AD spectrum. To embrace the necessary medical treatment effectively, it is crucial to pre-emptively identify the risk of getting the disease ahead at the earliest possible time. In order to diagnose AD, the brain's structural magnetic resonance imaging (sMRI) has been extensively exploited among other plausible neuroimaging \cite{apostolova2008,suk2015,zhao2021longitudinal} because it preserves macroscopic visual features \cite{suk2015} that can be extracted as essential AD-related biomarkers \cite{apostolova2008}. Despite the potential benefits of exploiting sMRIs, it is still considered a challenge because of the subtle and gradual progressive traits of AD with a diverse inter-subject progression rate \cite{davis2018} and high probability of getting entangled with the morphological changes of normal aging \cite{apostolova2008,sivera2019}.

We attempt to overcome the aforementioned concerns through AD progression modeling (ADPM), referring to any method that aims to perform diagnosis, monitoring, or prognosis on the risk of getting AD across its spectrum and/or across the course of time. A surging number of studies on sMRI-driven ADPM have been proposed and exhibited noteworthy achievements by means of machine learning (ML) methods, encompassing the computer-aided diagnoses field of research \cite{sukkar2012, nader2018, lee2019}. More recently, a family of deep learning models has brought a brighter light upon ADPM by unveiling distinguished outcomes compared to conventional ML methods \cite{zhang2020asurvey}. Notably, a variety of convolutional neural networks (CNNs) were tailored either on predetermined 2D slices of sMRI \cite{martinezmurcia2020}, whole-brain 3D sMRI \cite{basu2019,hosseiniasl2018,jun2021medicaltransformer}, or addressing the subtle morphological biomarkers on particular portions of sMRI (\ie,\ voxels, regions, or patches) \cite{samsuddin2020,zhu2021dualattention,lian2018hierarchical} in diverse approaches, including: by discovering meaningful latent features via  convolutional autoencoders (CAEs) \cite{basu2019,martinezmurcia2020,chadebec2022dataaugmentation}; by incorporating attention mechanisms \cite{yang2018,zhu2021dualattention}; or by utilizing adversarial strategies \cite{ravi2019, xia2019consistent, 0h2021lear, pan2021disease}. Those approaches addressed ADPM mostly for diagnostic purposes, including the clinical labels classification \cite{basu2019,hosseiniasl2018,jin2019,korolev2017,liu2020onthedesign}, cognitive scores \cite{martinezmurcia2020} or age \cite{cole2017} prediction, with the least concern over explainability of underlying AD progression. By conducting a longitudinal study, ADPM might discover more meaningful insights regarding the AD progression, such as by obtaining AD-related highlights on structural differences throughout years \cite{ocasio2021deep}, contrasting the rate of aging and AD progression \cite{Ouyang2022disentangling}, or simulating brains according to their ages \cite{zhao2021longitudinal, ravi2019, xia2019consistent}. However, a longitudinal study faces its own hindrances, namely the necessity to collect and utilize a vast amount of data subsequently, which is costly, time-consuming, and prone to inevitable missing data \cite{tabarestani2020}.

\begin{figure*}
\centering
\begin{subfigure}{0.63\textwidth}
\includegraphics[width=1.0\textwidth]{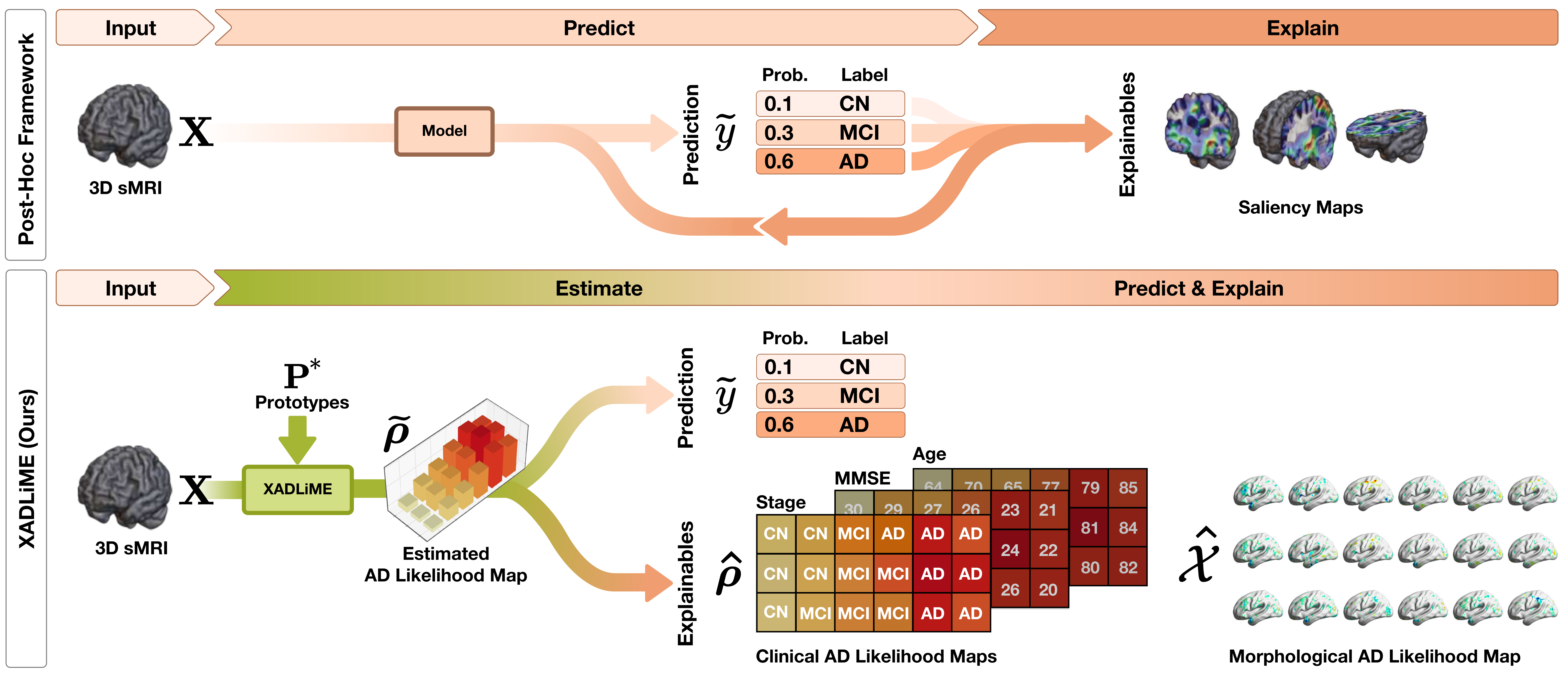}    \caption{Comparison of our proposed XADLiME to a post-hoc framework}
    \label{fig:framework_comparison}
\end{subfigure}
\hfill
\begin{subfigure}{0.35\textwidth}
\includegraphics[width=1.0\textwidth]{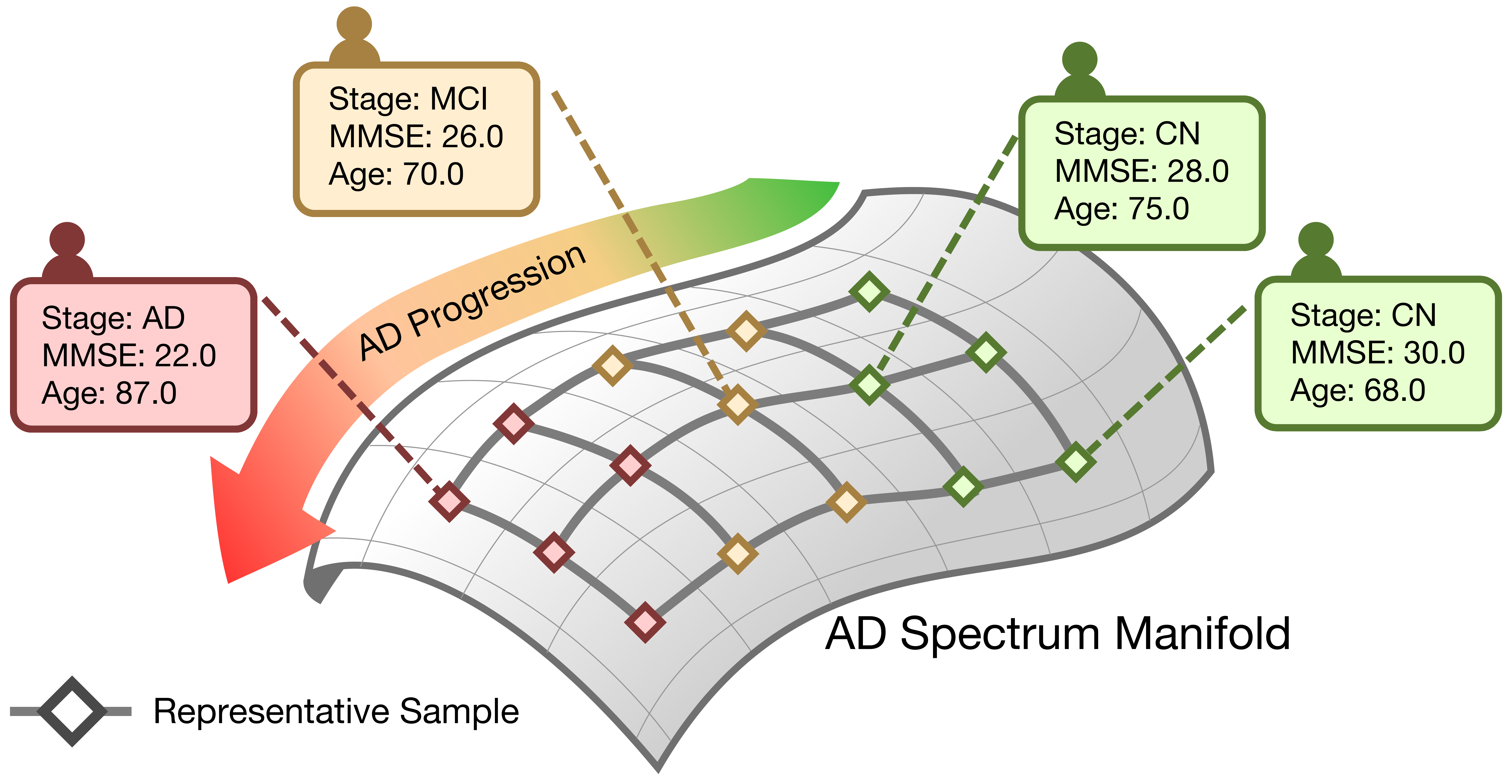}
\caption{A hypothetical AD spectrum manifold}
\label{fig:ad_spectrum_manifold}
\end{subfigure}
\caption{(a) Comparison of our approach (bottom) to a typical post-hoc framework (top) in an attempt to promote explainability. Our proposed XADLiME provides two complementary sets of explainable AD likelihood maps from the clinical as well as morphological perspective. (b) Illustration of a hypothetical AD spectrum manifold with a set of diamond markers denotes the representative samples, and each holds informative clinical states (\eg, clinical stage, cognitive score, and age), mimicking the AD progression pathways. A subtle progression can be achieved by maintaining the relationship across such representative samples. Furthermore, it is only natural and favorable to represent more than one sample per clinical stage to accommodate the prevailing diverse inter-subject variances.}
\label{fig:framework_overview}
\end{figure*}



Over the past few years, concerns regarding the explainability of artificial intelligence methods have been raised and are progressively growing as the standalone focus of research under the term `explainable artificial intelligence (XAI)', implying its significance, especially in domains that involve high-risk and life-threatening decisions (\eg,\  health and medical care) \cite{samek2019towards,jimenez2020drug,barnett2021case}. Among numerous efforts in endeavoring explainability (interchangeable with interpretability) under XAI, example-based explanations were introduced by picking a set of representative instances over the entire dataset to explain the inherent data distribution as well as the ML model's behavior under the hood \cite{molnar2022}. This direction is arguably more favorable as compared to \mbox{\emph{post-hoc}} approaches, which were frequently reported as being unreliable and can be misleading \cite{Rudin2019}. A post-hoc approach is a method that aims to uncover the explainability (\eg, visual explanation) of a certain trained model in drawing its decisions (which might not be readily interpretable by design) in a separate auxiliary step following the training phase \cite{Rudin2019, Barredo2020xai, li2018casebasedreasoning} as depicted in Fig. \ref{fig:framework_comparison} (top).

One branch of example-based explanations approach is the prototype-based network, which learns a set of class-specific prototypes (\ie, representative samples) \cite{molnar2022,li2018casebasedreasoning, chen2019looks, kim2021xprotonet,snell2017prototypical}. Here, each prototype is optimized to capture the resemblance to (at least) one sample \cite{li2018casebasedreasoning} or parts of a sample \cite{chen2019looks, kim2021xprotonet}, or to represent a single class by simply taking the mean over class-specific features \cite{snell2017prototypical} in the latent space. Interpretability (\ie, easily-interpreted concepts by humans) can be provided by either projecting each prototype to the closest features and further upsampling its similarity map \cite{chen2019looks, kim2021xprotonet}, or by utilizing the corresponding decoding network to transform it to the original dimension of input \cite{li2018casebasedreasoning}. It should be emphasized that initially,  typical prototype learning was not devised to consider the relationship or association among its prototype units, except that they merely penalized its proximity via diversity regularization \cite{ming2019steer, Trinh_2021_WACV}.

To this end, we argue that a satisfactory ADPM must meet the following criteria: (i) it intuitively considers the natural traits of AD (\eg, irreversibility and progressiveness); (ii) it provides explainability in the form of a set of representative features to reliably portray the progression facts of AD (clinically and/or morphologically); (iii) it yields adequate interpretability without sacrificing rigorous diagnostic performance. With these criteria in mind, we propose a framework for ADPM using a novel approach through eXplainable AD Likelihood Map Estimation (XADLiME) \footnote{Our implementation source code is available at \url{https://github.com/ku-milab/XADLiME}} over a whole-brain 3D sMRI scan via clinically-guided prototype learning. Specifically, we exploit composite clinical information (\eg,\ cognitive score, clinical stage, and age) to divulge a multitude of underlying facts about existing clinical conditions manifested in a brain sMRI scan of a subject. These features will be fed into our proposed AD-spectrum-aware prototypical embedding network (ADPEN) to establish a set of topological-aware prototype units onto a hypothetical AD spectrum manifold \mbox{(Fig. \ref{fig:ad_spectrum_manifold})}, with each unit holding a tuple of essential underlying clinical states. These sets are utilized as \emph{clinically-guided prototypes}. A \emph{``pseudo" AD likelihood map} can then be inferred by simply measuring the similarity of latent clinical features to those prototypes. By considering such a map as an enriched reference, we further employ an estimating network to estimate the AD likelihood map from the respective 3D sMRI, which serves as concise and discriminative features for the downstream clinical task (\ie, AD/MCI identification), while promoting interpretability from the clinical and morphological perspectives. We reaffirm the distinction of our approach in providing explainability compared to a typical aforementioned post-hoc framework as depicted in Fig. \ref{fig:framework_comparison} (bottom). 

Our main contributions in this work are as follows:

\begin{itemize}
    \item We propose XADLiME as a novel explainable predictive framework to tackle ADPM by estimating an explainable AD likelihood map aided by a set of clinically-guided prototypes, offering succinct and easily comprehensible interpretation while maintaining the robustness of the diagnostic performance for downstream tasks.
    \item Through the proposed ADPEN, we simultaneously discover a hypothetical AD progression manifold portrayed with a set of topological-aware prototypes in an unsupervised manner by exploiting auxiliary clinical embedded features, eliminating the necessity to predetermine the number of class-specific prototypes. 
    \item We promote explainability from clinical and morphological perspectives by exploiting the established clinically-guided prototypes. From the clinical perspective, given that ADPEN can decode each prototype as a representative clinical state, we merge its values with the estimated AD likelihood map over a brain sMRI to highlight the current clinical states of a subject. Furthermore, we devised an algorithm to retrieve a set of prototypical brains to simulate the brain changes of an unseen sMRI scan, providing explainability from the morphological perspective.
    \item We demonstrate extensive interpretability analysis for the potential utilization of XADLiME in terms of longitudinal clinical applications for the estimated  AD likelihood map on several progression study cases by utilizing longitudinal data.
\end{itemize}
We validated the effectiveness of the proposed method by conducting exhaustive experiments and analysis using 3D brain sMRI scans on the Alzheimer’s Disease Neuroimaging Initiative (ADNI)\footnote{Publicly available at \url{http://adni.loni.usc.edu/}} dataset \cite{mueller2005}. 

\begin{figure*}[!ht]
    \centering
    \includegraphics[width=0.7\textwidth]{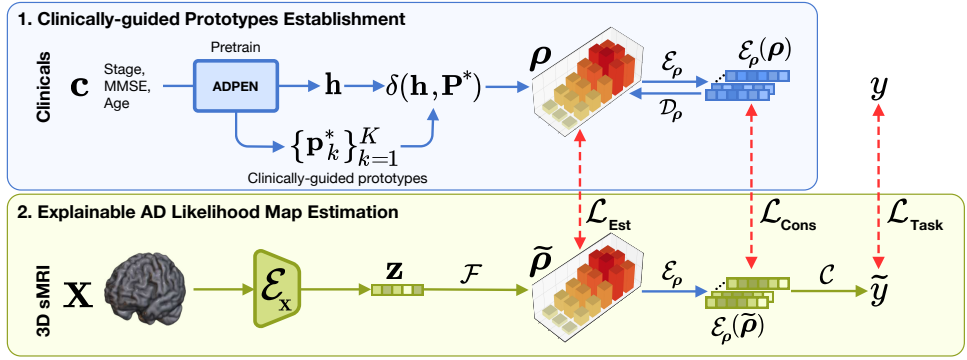}
    \caption{Overview of XADLiME in estimating the AD likelihood map $\widetilde{\vrho}$ over a 3D sMRI. The estimation is aided by a ``pseudo" likelihood map $\vrho$ by measuring the similarity of latent clinical features $\rvh$ with the clinically-guided prototypes $\rmP^*$. A predicted label $\widetilde{y}$ is inferred from estimated map $\widetilde{\vrho}$ via a joint encoding and task-dependent prediction network.}
    \label{fig:xadlime_framework}
\end{figure*}

\section{Related Works}

Prototype learning for addressing ADPM using sMRIs is still relatively limited and under-explored. To name a few examples, the generalized metric learning vector quantization (GMLVQ) framework was proposed in \cite{GIORGIO2020102199} over brain imaging modalities, including sMRI scans, by iteratively learning a prototype per class along with its decision boundaries. DPTree \cite{zhang2021dptree} employed prototype and features-ordering learning over multi-modalities of structure and functional MRI jointly. Those methods merely utilized one prototype per class, neglecting substantial inter-subject variances for each class. In addition, \cite{zhang2021dptree} required an auxiliary tool to visualize the learned AD spectrum progression in the form of a tree structure. Furthermore, \cite{ortiz2013lvq} utilized a similar LVQ to build ADPM discriminative features upon probability distribution of voxel intensities over 3D sMRIs, by previously segmenting it into the tissue distribution profiles through a set of prototype vectors. For drawing the final verdicts, they relied upon the subsequent training of an ML-based classifier. Furthermore, \cite{mohammadjafari2021using} re-utilized the ProtoPNet \cite{chen2019looks} in learning a set of predefined class-specific prototypes over 2D sMRI scans, highlighting the AD-related biomarkers in certain parts of the input image. Despite the use of multiple prototypes per class, the underlying relationship among prototypes does not exist in their work. Finally, through our preliminary work in \cite{mulyadi2021protobrainmaps}, we have attempted to tackle ADPM via a prototype learning approach where we explicitly devised the relations among prototypes as a chain by restraining the distances between two adjacent prototypes. We acquired an insightful glimpse of a hypothetical AD progression manifold in terms of the brain morphological changes of 3D sMRI by establishing prototypes over the clinical and brain imaging features. However, as we devised two distinct sets of prototypes in both feature spaces (\ie, clinical and brain imaging), it was rather challenging to impose its alignment owing to the large gap in dimensionalities. Hence, in this work, we resort to a novel approach in addressing the ADPM by establishing a single set of prototypes in a learned clinical embedding space and regarding those as an enriched reference for discovering AD progression.

\section{Method}

We denote the neuroimaging dataset with $N$ number of subjects as  $\{\rmX_n, \rvy_n, s_n, a_n\}_{n=1}^N$, where $\rmX_n \withdim{ C \times W \times H \times D}$ denotes the 3D sMRI scan of $n$-th subject with $C, W, H,$ and $D$ represents its channel, width, height, and depth, accordingly. Each scan is coupled with corresponding clinical measurements describing the brain of the $n$-th subject, including: clinical stage label $\rvy_n \in \{0,1\}^L$, cognitive score $s_n \withdim{}$, and age $a_n \withdim{}$, with $L$ denoting the number of classes. Hereafter, we omit the subscript $(n)$ for clarity. 

As depicted in Fig. \ref{fig:xadlime_framework}, we present the main ideas of our XADLiME framework, which is comprised of two subsequent essential streams: (i) the establishment of clinically-guided prototypes over the latent clinical features in an AD spectrum manifold via our proposed ADPEN; (ii) the estimation of explainable AD likelihood map over 3D sMRI scan via an estimating network assisted by well-interpolated clinically-guided prototypes.

\subsection{Clinically-guided Prototypes Establishment} 

During this initial step, we aim to establish $K$ number of prototypes $\rmP = \{\rvp_k \}_{k=1}^K$ over the latent clinical features $\rvh$ in a hypothetical AD spectrum manifold. In order to discover such manifold, we devise ADPEN, which is built upon a VAE as the family of the deep generative model along with two key components, namely: (i) an AD-spectrum-aware ordering loss to impose the features in comprehending the AD progression and (ii) a SOM module to establish a set of topological-aware prototypes adequately. 

We define the composite clinical features $\rvc = [\rvy \circ s \circ a]$, with $\rvc \withdim{L+2}$ as the concatenation of the clinical label, cognitive score and age of a subject, assuming that it was influenced by latent variable $\rvh \withdim{M}$ under $p_\theta(\rvc \vert \rvh)$, with $M$ denoting its dimensionality. The latent features $\rvh$ is drawn from the prior distribution $p_\theta(\rvh)$, reflecting the underlying AD-related health conditions of a subject. Such features could be inferred from the clinical features under the true posterior distribution $p_\theta(\rvh \vert \rvc)$. Through a VAE, we approximate it by introducing $q_\phi(\rvh \vert \rvc)$ using Gaussian $\gN(\vmu_\rvh, \textrm{diag}(\vsigma_\rvh^2))$, with means $\vmu_\rvh \withdim{M}$ and variances $\vsigma_\rvh^2 \withdim{M}$ that are inferred from an encoding network $\gE_\rvc$ with parameter $\phi$ as
\begin{equation}
    \vmu_\rvh = \gE_\rvc(\rvc; \phi), \quad 
    \log \vsigma_\rvh^2 = \gE_\rvc(\rvc; \phi).
\end{equation}
To train VAEs via a regular stochastic gradient methods, we employ reparameterization trick \cite{kingma2014} as $\rvh = \vmu_\rvh + \vsigma_\rvh \odot \vepsilon$, where $\odot$ denotes the element-wise multiplication operator, while  $\vepsilon \sim \gN(\vzero, \rmI)$ denotes the sampling from a Gaussian with zero mean and unit variance. The reconstructed $\widetilde{\rvc}$ can then be derived from decoder $\gD_\rvc$ with parameter $\theta$ via $\widetilde{\rvc} = \gD_\rvc(\rvh; \theta)$. Eventually, we train VAEs by maximizing the variational evidence lower bound via reconstruction loss and Kullback–Leibler (KL) divergence as

\begin{equation}
    \gL_{\textrm{VAE}} = \sum^N_{n=1} \mathbb{E}_{q_\phi(\rvh_n \vert \rvc_n)}[\log p_\theta(\rvc_n \vert \rvh_n)] 
 - \textrm{KL}(q_\phi(\rvh_n \vert \rvc_n) \lVert p_\theta(\rvh_n)).
\label{eq:loss_vae}
\end{equation}

In imposing the AD-spectrum-aware ordering to comprehend the AD progression pathway, we utilize an analogous approach to the longitudinal self-supervised learning (LSSL) \cite{zhao2021longitudinal}. However, such LSSL imposes the order of the latent features across a series of longitudinal sMRI scans in chronological order according to its acquisition time. In contrast, as we utilize the baseline (first-visit) sMRI scans, we devise our ADPEN to comply with and comprehend the natural order of the AD spectrum by drawing paired samples of two distinct samples at subsequent stages (\eg, CN$\rightarrow$MCI, MCI$\rightarrow$AD). Given latent features $\rvh^l$ with stage $l$, we draw a sample with the corresponding subsequent stage in the AD spectrum $\bar{\rvh}^{l+1}$ randomly (\eg, CN as stage $l$ and MCI as stage $l+1$), which shall be enforced to obey the ordering through Eq. (\ref{eq:loss_order}) by using a trainable layer  $\gO$ with parameter $\psi$.  
\begin{equation}
    \gL_{\textrm{Order}} = \sum_{n=1, j \neq n}^{N}  \frac{\gO(\rvh_n^l; \psi)-\gO(\bar{\rvh}_{j}^{l+1};\psi)}{\distance{\rvh_n^l - \bar{\rvh}_{j}^{l+1}}_2}
    \label{eq:loss_order}
\end{equation}

Furthermore, we then utilize the SOM \cite{kohonen1990} as a family of neural networks that possess a prominent ability to organize the internal representations (\ie,\ prototype units) of numerous features topologically and has proven its compatibility with various deep models \cite{forest2019,mulyadi2021protobrainmaps}. We nominate the SOM-based clustering layer as it offers a topological arrangement (\eg,\ 1D chain-like, 2D grid-like, and so on) over a set of prototypes that benefits ADPM. This 1D topological arrangement of prototypes can be imagined as a chain (\ie, each unit has two adjacent prototypes, excluding the ones at the very end) that spreads over the clusters of the AD spectrum. Likewise, we impose the grid-like topological relations among prototypes for the 2D arrangement (with four adjacent units excluding the outskirts) and further add its hierarchical layers (\ie, depths) for the 3D topology.

We initialize a SOM module on top of VAE's encoding network with $K$ number of prototypes as $\rmP = \{\rvp_k \}_{k=1}^K$, with the $k$-th prototype unit denoted as $\rvp_k \withdim{M}$. Those $K$ prototypes shall be organized in a certain way to abide the predefined topological arrangement. Note that we initialize $M$ as the prototype dimensionality, which is identical to the latent clinical features $\rvh$. Given the latent clinical features $\rvh$ and those prototypes $\rmP$, a typical prototype network is trained by means of distance function $\delta$ (\eg, squared Euclidean distance) as
\begin{equation}
    \delta(\rvh, \rvp_k) = \distance{\rvh - \rvp_k}_2^2 \quad \forall \ k \in \{1, \ldots, K\}.
\end{equation}
We impose the framework to establish those prototypes as adequately as possible to the clusters of latent clinical features. In contrast, SOM requires a distinctive learning mechanism mainly owing to maintaining the inherent topological constraint among prototypes. For this, we begin by selecting the most closest prototype via best matching unit (BMU) index $\beta$ through an assignment $\beta = \argmin_k \delta(\rvh, \rvp_k)$. However, this assignment operation is impractical to be differentiated, hampering our aim to unify the SOM with a more contemporary deep model, such as VAE in our case. Inspired by \cite{forest2019}, to fuse the learning mechanism of conventional SOM with a deep generative model seamlessly, we introduce $\vomega \withdim{K}$ for weighing the distances $\delta(\rvh, \rmP)$ in updating the prototypes by considering the vicinity-decayed topological relation as
\begin{align}
    \vDelta &= \gT(\beta, {\rmP}),
\end{align}
\begin{align}
    \vomega = \exp \bigg( -\frac{\vDelta^2}{2\Gamma^2} \bigg), \ \text{where} \ \Gamma = \Gamma_{\textrm{max}}\left(\frac{\Gamma_{\textrm{min}}}{\Gamma_{\textrm{max}}}\right)^{t/T}.
\end{align}
Here, $\gT$ denotes the topological distance metric (\eg, Manhattan) with $\vDelta \withdim{K}$ as its measured distances of BMU index $\beta$ to the rest of prototypes' indexes in $\rmP$. We consider the vicinity in updating the prototypes topologically via a Gaussian neighborhood such that the effect of learning shall be proportional to the distance $\vDelta$ of the neighboring prototypes \wrt\ the BMU (\ie,\ prototype units that are further away from BMU should be less updated and vice versa). We control this vicinity through $\Gamma \withdim{}$ as the decaying radius based on the parameter $\Gamma_{\textrm{min}}$ and $\Gamma_{\textrm{max}}$, while $t$ and $T$ denote the current iteration and total number of iterations, respectively. By employing $\vomega$, we update a large portion of adjacent prototypes \wrt\ to the BMU prototype at the beginning of iterations while gradually shrinking its updates' strength and coverage topologically across iterations. The SOM module can then be optimized as 
\begin{equation}
    \gL_{\textrm{SOM}} = \sum_{n=1}^N \sum_{k=1}^K  \vomega_n \odot \delta(\rvh_n, \rvp_k).
    \label{eq:loss_som}
\end{equation}
We jointly optimize all trainable parameters $\{\phi, \theta, \psi, \rmP\}$ of ADPEN in Eq. (\ref{eq:loss_adpen}) with hyperparameter $\lambda_1$ for controlling the learning capability of SOM module. To this end, our proposed ADPEN is optimized to simultaneously reveal, reconstruct, and order the latent clinical features while establishing the prototype units in a latent embedding space. Upon optimizing $\gL_{\textrm{ADPEN}}$, we acquire a set of well-established clinically-guided prototypes $\rmP^*$ that can be utilized as an enriched reference for estimating the explainable AD likelihood map.
\begin{equation}
    \gL_{\textrm{ADPEN}} = \gL_{\textrm{VAE}}  + \gL_{\textrm{Order}} + \lambda_1 \gL_{\textrm{SOM}}
    \label{eq:loss_adpen}
\end{equation}


\subsection{Explainable AD Likelihood Map Estimation} 
We further utilize the well-established clinically-guided prototypes $\rmP^*$ by measuring the similarities against latent features $\rvh$ of the clinical information $\rvc$, leading to a ``pseudo" likelihood map $\vrho \withdim{K}$ which is inferred as
\begin{equation}
    \rho_k = \frac{\textrm{exp} (-\delta(\rvh, \rvp_k^*)/\gamma)}{\sum_j \textrm{exp} (-\delta(\rvh, \rvp_j^*)/\gamma)} \quad \forall \ k \in \{1, \ldots, K\},
\end{equation}
where $\gamma$ denotes the distilling temperature parameter to regulate softer probabilities \cite{hinton2015distilling}. In practice, we swap such a parameter by accounting for inherent variance \cite{kim2021distilling}. We additionally apply the min-max normalization over such pseudo-likelihood map $\vrho$ for a better training process. 

Intuitively, we obtain such a likelihood map as a set of scores in highlighting which clinically-guided prototype most resembles the clinical information in question. The highest score indicates the closest prototype, leading to the most representative regarding the underlying clinical states. Furthermore, the soft probabilities operation and the inbuilt topological arrangement between prototypes will guarantee that the nearby prototypes exhibit a substantial resemblance to the highest one, capturing the facts of the subtle progression of AD. We end up with such a likelihood map as an easily-interpreted heatmap to a set of decoded clinical information over the prototypes through $\{\gD_\rvc(\rvp_k^*)\}_{k=1}^K$.

To capture inherent progression association over the adjacent values in likelihood map $\vrho$, we employ a light CAE network that is pre-trained for reconstructing the $\vrho$ that is optimized via $\sum_{n=1}^N \distance{\vrho_n - \gD_{\vrho}(\gE_{\vrho}(\vrho_n))}_2^2$. Afterward, we require its encoding network $\gE_{\vrho}$ to be re-utilized as a supplemental consistency mechanism in evaluating the inherent agreement between the pseudo-likelihood map and the estimated likelihood map over 3D sMRI.

To this end, we denote the estimated likelihood map from the brain imaging as $\widetilde{\vrho}$, which is devised in a way that it properly comprehends the pseudo-likelihood map $\vrho$ as its reference. Note that the dimensionality and topological arrangement of $\vrho$ and $\widetilde{\vrho}$ shall be identically initialized. Specifically, we infer the latent features $\rvz \withdim{\ell}$ over sMRI scan through a feature extractor network $\gE_\rmX$ with a trainable parameter $\varphi$ as $\rvz = \gE_\rmX(\rmX;\varphi)$, reflecting these features as the underlying clinical conditions manifested in the brain of a subject. We could then estimate the AD likelihood map  $\widetilde{\vrho} \withdim{K}$ via a FFNs $\gF$ as
\begin{equation}
    \widetilde{\vrho} = \sigma(\gF(\rvz; \Omega)),
\end{equation}
with $\Omega$ denoting its trainable parameter and $\sigma$ denoting a sigmoid activation function. For this likelihood map estimation task, we train the framework through $\gL_{\textrm{Est}}$ by evaluating the element-wise estimation error between two paired maps. We also devise an auxiliary objective function $\gL_\textrm{Cons}$ to consider the consistency of the inherent relationship between two likelihood maps topologically via $\gE_{\vrho}$. Both $\gL_{\textrm{Est}}$ and $\gL_\textrm{Cons}$ are evaluated using MAE and MSE as

\begin{equation}
    \gL_\textrm{Est}(\vrho,\widetilde{\vrho}) = \distance{\vrho - \widetilde{\vrho} }_1 + \distance{\vrho - \widetilde{\vrho} }_2^2,
\end{equation}
\begin{equation}
    \gL_\textrm{Cons}(\vrho, \widetilde{\vrho}) =  \distance{\gE_{\vrho}(\vrho) - \gE_{\vrho}(\widetilde{\vrho}) }_1 + \distance{\gE_{\vrho}(\vrho) - \gE_{\vrho}(\widetilde{\vrho}) }_2^2.
\end{equation}

Having estimated the well-represented likelihood map $\widetilde{\vrho}$, we could then employ it as the subsidiary features over sMRI scan for diagnostic purposes. We formulate the loss function $\gL_{\textrm{Task}}$ for the downstream diagnostic task by estimating the predicted label $\widetilde{\rvy}$ via a network $\gC$ with a trainable parameter $\Phi$ as
\begin{equation}
    \gL_{\textrm{Task}}(\rvy, \widetilde{\rvy}) = - \sum_{l=1}^L y^l \log \widetilde{y}^l, \ \text{with} \  \widetilde{\rvy} = \gC(\gE_{\vrho}(\widetilde{\vrho}); \Phi).
\end{equation}
Note that $\gL_{\textrm{Task}}$ denotes the task-dependent loss, which is defined as the cross-entropy loss to optimize the overall networks for classification task (\ie, clinical stage prediction). Without losing generalization, we also devise $\gL_{\textrm{Task}}$ for the cognitive score and age prediction tasks by optimizing the network to minimize the MSE between ground truth and its prediction value instead. Finally, we optimize the integrated set of trainable parameters $\{ \varphi, \Omega, \Phi\}$ in Eq. (\ref{eq:loss_estimating}) through the objective functions on estimation and topological relation for evaluating the agreement between two entities of maps, as well as the downstream clinical task, jointly.
\begin{equation}
    \gL_{\textrm{Total}} = \frac{1}{N} \sum_{n=1}^N \gL_\textrm{Est}(\vrho_n,\widetilde{\vrho}_n) + \lambda_2\gL_\textrm{Cons}(\vrho_n, \widetilde{\vrho}_n) + \lambda_3\gL_{\textrm{Task}}(\rvy_n, \widetilde{\rvy}_n)
    \label{eq:loss_estimating}
\end{equation}
We define $\lambda_2$ and $\lambda_3$ as the hyperparameters for weighing the consistency and task-dependent losses, respectively. By optimizing $\gL_{\textrm{Total}}$, we could then obtain a comprehensive estimated likelihood map $\widetilde{\vrho}$ of a subject given its brain sMRI.

As an extra benefit, our proposed XADLiME offers thorough composite explainable AD likelihood maps from \emph{clinical} as well as \emph{morphological perspectives} as illustrated in Fig. \ref{fig:framework_comparison} (bottom). 

\subsubsection{Clinical AD Likelihood Maps}
From a clinical perspective, we desire to provide comprehensive interpretability in explaining the given 3D sMRI with regard to \emph{their clinical status in terms of the AD spectrum}. Here, we could assess the current clinical progression stage in the stretch of the AD spectrum by simply merging it with a set of decoded clinical states out of the clinically-guided prototypes through $\{\gD_\rvc(\rvp_k^*; \theta^*)\}_{k=1}^K$ with the estimated map $\widetilde{\vrho}$ fulfilling the role as the heatmap to those set of clinical states. Henceforth, we denote such an explainable AD likelihood map from the clinical perspective as $\hat{\vrho} \withdim{K}$ comprised of a tuple over a similarity score and the decoded prototypical clinical information as $\hat{\vrho}_k = [\widetilde{\rho}_k, \gD_\rvc(\rvp_k^*; \theta^*)] \ \forall \ k \in \{1,\ldots,K\}$.

\subsubsection{Morphological AD Likelihood Map}
We also provide in-depth explainability of the AD likelihood map by inspecting \emph{the changes upon the brain morphology}. As we traverse along the AD pathological pathway, we expect to observe AD-related biomarkers (\ie, atrophy, cortical thinning, and so on). For this purpose, once again, we exploit the clinically-guided prototypes $\rmP^*$ as the reference. Specifically, we obtain a set of prototypical brains over the training samples through $\argmin_n \delta(\rvh_n, \rvp^*_k)$ for every $k$-th prototype unit on the entire $\{1, \ldots, K\}$ set of learned prototypes $\rmP^*$. By doing so, we shall acquire the closest clinical information, leading to obtaining the respective sMRI scan eventually. We denote such a set of prototypical brains as $\gX = \{\bar{\rmX}_k\}_{k=1}^K$. Assuming that those sets of sMRI scans were specifically pre-processed using a non-linear registration mechanism, the brain regions shall be aligned into a common reference. We then could simulate the morphological changes by subtracting the test sample $\rmX$ from selected prototypical brains in $\gX$ via $\hat{\gX} = \{\rmX - \bar{\rmX}_k \}_{k=1}^K$. 

\section{Experimental Settings}

\subsection{Dataset}
We evaluated our XADLiME framework on ADNI \cite{mueller2005}, a publicly available AD dataset. Each 1.5/3T T1-weighted 3D brain sMRI scan sample in such a cohort was coupled with the corresponding clinical information, comprising a clinical label, a mini-mental state examination (MMSE) score, and age. If missing clinical information was encountered, we excluded the respective 3D brain sMRI as a training sample. We utilized two study phases of ADNI-1 and ADNI-2, which resulted in a cohort comprising 1,540 first-visit subjects in total. This cohort was comprised of 28.12\% samples labeled as CN, 48.57\% as MCI, and the remaining 23.31\% as AD. In addition, there existed two sub-intermediate stages of MCI towards the AD stage, which were stable MCI (sMCI) with 66.44\% and progressive MCI (pMCI) with 33.56\% over the total MCI samples. We considered the pMCI subjects as the ones that eventually progress to the AD stage within the following 24 months span \cite{zhou2020svm}. We split the cohort into training, validation, and testing sets using stratified five-fold cross-validation. Furthermore, we applied a series of pre-processing procedures and normalization upon 3D sMRI scans \cite{jenkinson2012,isensee2019} as well as its clinical information described in detail in Appendix A.1.

\subsection{Model architecture and training}
We devised ADPEN with three layers of FFNs for the encoding network of VAEs with the number of hidden units of $\{10, 16, 8, 3\}$, while $M=3$  for latent features $\rvh \withdim{M}$. The subsequent decoding networks were utilized using an identical number of hidden units in reversed order. ReLU was employed as the intermediate activation function in both networks, with a softmax for the last layer for the clinical stage. As for the SOM module, we conducted an ablation study where we set the number of prototypes $K=\{64, 100\}$ with several settings of topological arrangements, namely 1D, 2D, and 3D topology. We jointly trained this ADPEN using stochastic gradient descent with an Adam optimizer using the following parameters: 1,000 epochs, four mini-batches, and a learning rate of 0.001. We fine-tuned the SOM module with fixed VAEs for an additional 500 epochs to encourage adequate prototype coverage. 

Through XADLiME, we re-utilized the clinically-guided prototypes to aid in estimating AD likelihood maps over 3D sMRIs. Specifically, we designated a CNN-based network for feature extractor $\gE_{\rmX}$ over the 3D sMRI with five layers of convolutional blocks, each comprised of two sets of convolutional layer-batch normalization-ReLU, where the first set plays the role of further reducing the dimensionality of the feature maps. We initialized the first layer's output channel to be $32$, and further scaled it by a factor of two throughout the layers. As a result, we acquired the flattened latent features of sMRI $\rvz$ with the dimension of 512. We then flattened out such features and fed them onto an estimating network $\gF$ as FFNs with a hidden layer with 256 hidden units and output the likelihood map $\vrho \withdim{K}$, where $K$ indicated the settings of ADPEN for the number of prototypes. Finally, the task-dependant diagnostic network $\gC$ was also an FFN with 64 hidden units estimating class prediction probabilities with a softmax on its last layer for the classification or sigmoid in the case of MMSE and age prediction tasks. We optimized these networks using an Adam optimizer with 200 epochs, ten mini-batches, and a learning rate of 0.0001 and 0.01 for clinical stage classification and regression tasks, respectively, with an exponential learning rate decay of 0.98 every ten epochs. We arranged the comparable setting and parameters for each baseline for a fair comparison. See Appendix B and C for details of each network and training hyperparameters, respectively.

\subsection{Downstream Clinical Tasks and Evaluation Metrics}
We assessed the performance of the proposed methods as well as baselines in performing several downstream clinical tasks, namely: i) clinical stage classification, ii) cognitive score prediction and iii) age prediction. 

For the clinical stage classification, we measured the performance of models via the area under the curve (AUC) of receiver operating characteristics (ROC) score, balanced accuracy (Bal. ACC), and F1-score. These metrics were measured for all classification scenarios, including i) CN/MCI; ii) MCI/AD; iii) CN/AD; iv) sMCI/pMCI, and; iv) CN/MCI/AD. For the multiclass classification scenario, we calculated one-vs-rest (OVR) multiclass of AUC score and utilized weighted multiclass F1-score\footnote{Using the available method at \url{http://scikit-learn.org}}. We selected the best model across the training epochs according to the highest model's accuracy on the validation set and further reported the average and its standard deviation over the test sets from the entire five folds.

We normalized the original MMSE score -- discreetly ranging from $0$ (severe) up to $30$ (healthy) -- to have a spread between $[0, 1]$. As such, we considered the prediction of the cognitive score as a regression task. We evaluated the models' performances using the root mean squared error (RMSE) and coefficient of determination ($R^2$), where we picked the best models based on the lowest validation MSE. Finally, we utilized a similar approach for the age prediction task in terms of its normalization and evaluation metrics. However, we set the maximum age to be $100$ for the min-max normalization setup. Unlike the previous tasks, we solely employed healthy brain samples in the cohort for this age prediction task to avert mixing it with the deviated AD-manifested brains.

\subsection{Baselines} 
To validate the effectiveness of our proposed framework for ADPM, we selected several comparable methods that were proposed for AD diagnosis by utilizing 3D whole-brain sMRI scans, including Jin \etal\ \cite{jin2019}, Korolev \etal\ \cite{korolev2017}, and Liu \etal\ \cite{liu2020onthedesign}. In a nutshell, Jin \etal\ \cite{jin2019} employed ResNet and further enhanced it using the attention mechanism to improve the classification performance while identifying AD-related regions simultaneously. We incorporated VoXCNN proposed by Korolev \etal\ \cite{korolev2017}, which took inspiration from VGG by first employing a series of volumetric convolutional blocks to extract the meaningful and discriminative features before feeding them onto several layers of FFNs as the final classifier. Finally, Liu \etal\ \cite{liu2020onthedesign} proposed a convolutional-based model, which utilizes instance normalization, small-sized kernels in the early layers, and a widening mechanism. In addition, we also included vanilla ResNet-18\ \cite{he2016} as one of the baselines as it is considered the most commonly exploited architecture for the backbone network for natural and medical images.

\section{Experimental Results}

\subsection{Downstream Clinical Tasks}

\begin{table}[]
 \caption{Clinical stage classification performances on the ADNI dataset.}
\label{table:classification_adni}
\centering
\scalebox{0.88}{
\begin{tabular}{clccc}
\toprule
\textbf{Scenario} & \multicolumn{1}{c}{\textbf{Models}} & \multicolumn{1}{c}{\textbf{AUC}} & \multicolumn{1}{c}{\textbf{Bal. ACC}} & \multicolumn{1}{c}{\textbf{F1-Score}} \\ \midrule
\multirow{5}{*}{\rotatebox[origin=c]{90}{CN/AD}} & ResNet-18\ \cite{he2016} & 0.9282±0.0364 & 0.8781±0.0448 &	0.8660±0.0493 \\
 & Jin \etal\ \cite{jin2019} & 0.9391±0.0320 & 0.8937±0.0324 &	0.8829±0.0358 \\
 & Korolev \etal\ \cite{korolev2017} & 0.9341±0.0272 & 0.8885±0.0397	& 0.8755±0.0441 \\
 & Liu \etal\ \cite{liu2020onthedesign} & 0.9438±0.0247 & 0.8971±0.0485	& 0.8855±0.0561 \\
 & \textbf{XADLiME} & \textbf{0.9492±0.0330} & \textbf{0.9183±0.0320} &	\textbf{0.9102±0.0356} \\ 
 \midrule
\multirow{5}{*}{\rotatebox[origin=c]{90}{CN/MCI}} & ResNet-18\ \cite{he2016} & 0.7154±0.0454	& 0.6430±0.0295 &	0.7682±0.0214 \\
 & Jin \etal\ \cite{jin2019} & 0.7140±0.0308 & 0.6437±0.0193 &	0.7523±0.0206 \\
 & Korolev \etal\ \cite{korolev2017} & 0.7305±0.0153 & 0.6511±0.0369 &	0.7432±0.0439 \\
 & Liu \etal\ \cite{liu2020onthedesign} & 0.7461±0.0545 & 0.6621±0.0457	& 0.7475±0.0815 \\
 & \textbf{XADLiME} & \textbf{0.7517±0.0355} & \textbf{0.6905±0.0439} &	\textbf{0.7747±0.0458} \\
 \midrule
\multirow{5}{*}{\rotatebox[origin=c]{90}{MCI/AD}} & ResNet-18\ \cite{he2016} & 0.7292±0.0425 & 0.6547±0.0449 &	0.5175±0.0705 \\
 & Jin \etal\ \cite{jin2019} & 0.7599±0.0681 & 0.6803±0.0418 &	0.5597±0.0534 \\
 & Korolev \etal\ \cite{korolev2017} & 0.7273±0.0648 & 0.6693±0.0904 &	0.5356±0.1329\\
 & Liu \etal\ \cite{liu2020onthedesign} & 0.7519±0.0384 & 0.6533±0.0206 &	0.5032±0.0459 \\
 & \textbf{XADLiME} & \textbf{0.7758±0.0584} & \textbf{0.7221±0.0530}	& \textbf{0.6173±0.0742} \\ 
 \midrule
 \multirow{5}{*}{\rotatebox[origin=c]{90}{sMCI/pMCI}} & ResNet-18\ \cite{he2016} & 0.6988±0.0206	& 0.6398±0.0192 &	0.5037±0.0763 \\
 & Jin \etal\ \cite{jin2019} & 0.6851±0.0647 &  0.6421±0.0194 & 0.5099±0.0406 \\
 & Korolev \etal\ \cite{korolev2017} & 0.7215±0.0106	& 0.6602±0.0368 &	0.5341±0.0648 \\
 & Liu \etal\ \cite{liu2020onthedesign} & 0.7329±0.0510 & 0.6584±0.0330 &	0.5309±0.0491 \\
 & \textbf{XADLiME} & \textbf{0.7567±0.0582} & \textbf{0.6785±0.0445} &	\textbf{0.5576±0.0732} \\ 
\midrule
  \multirow{5}{*}{\rotatebox[origin=c]{90}{CN/MCI/AD}} & ResNet-18\ \cite{he2016} & 0.7433±0.0596 & 0.6038±0.0433 &	0.5745±0.0501 \\
 & Jin \etal\ \cite{jin2019} & 0.7465±0.0488 & 0.5887±0.0318 & 0.5651±0.0437 \\
 & Korolev \etal\ \cite{korolev2017} & 0.7493±0.0428 & 0.5893±0.0229 &	0.5700±0.0298\\
 & Liu \etal\ \cite{liu2020onthedesign} & \textbf{0.7685±0.0225} & 0.5974±0.0444  & 0.5828±0.0424 \\
 & \textbf{XADLiME} & 0.7652±0.0386	& \textbf{0.6316±0.0519} & \textbf{0.6103±0.0573} \\ \\
 \bottomrule
\end{tabular}
}
\end{table}

As the establishment of clinically-guided prototypes plays a critical role in adequately guiding the estimation of the AD likelihood map, which eventually influences the downstream prediction network in concluding the final clinical outcomes, we conducted ablation studies over this establishment, as described in Appendix D.1. Specifically, we kept the number of prototypes on $K = \{64, 100\}$ and organized the set into 1D, 2D, and 3D topological arrangements. Overall, we discovered that the settings using 2D topology consistently achieved better classification performance compared to their 1D and 3D topological counterparts. Furthermore, the proposed framework achieved its majority highest performance with a 2D topological arrangement ($5 \times 20$) over the $K=100$ total number of prototypes. Here, we argue that a 2D grid-like topology with an adequate number of prototypes exhibited a beneficial mechanism in providing a series of prototypes across the AD spectrum while simultaneously accommodating the intra-stage variations as meaningful information to estimate the AD likelihood map and to draw better clinical outcomes. Thus, we utilized and reported the results from the discovered optimum setting for the remaining experiments and analysis, \ie, 100 prototypes with a 2D topology.

To evaluate the effectiveness of our proposed framework compared to the baselines, we conducted a series of downstream clinical tasks, including clinical stage classification, MMSE, and age prediction. The experimental results of comparing our proposed XADLiME to the baselines in performing clinical classification tasks as the primary purpose of ADPM for overall classification scenarios are presented in Table \ref{table:classification_adni}. Our proposed method considerably achieved higher classification performances in almost all scenarios. The performance of our XADLiME was slightly lower for the CN/MCI/AD three-class classification scenario than Liu \etal\ \cite{liu2020onthedesign} in terms of AUC with a relatively small gap but achieved significantly better (balanced) accuracy and F1-score. Meanwhile, the sMCI/pMCI classification is considered a difficult task because of the subtle morphological changes between these sub-stages of MCI. Despite that, our XADLiME substantially performed better than the competing baselines by virtue of the estimated AD likelihood map. By observing these results, we verified that the estimated AD likelihood map successfully accomplished its role as the substitute over the 3D sMRIs and provided discriminative features for drawing more accurate clinical verdicts.

We also evaluated the proposed method and baselines for the regression tasks. The experimental results for MMSE and age prediction are summarized in Appendix D.2. Likewise, we observed that  XADLiME attained the best performance in RMSE and $R^2$ for the MMSE as well as age prediction tasks. These results exhibit the distinguished potential of XADLiME for addressing ADPM in performing the downstream clinical tasks. As the proposed XADLiME achieved remarkable diagnostic performance, it accomplished one of the key criteria for being an ideal ADPM approach.

\subsection{Explainability Analysis}
As an explainable predictive framework, our XADLiME offers extra benefits in addition to maintaining rigorous performance over diverse downstream clinical tasks. This section is dedicated to further analyzing XADLiME by providing a couple of interpretations over (i)  the established clinically-guided prototypes in the AD spectrum manifold and (ii) the estimated AD likelihood map from a 3D sMRI scan. Furthermore, we jointly merge these two beneficial pieces of information, leading to an explainable AD likelihood map that provides concise highlights from the clinical perspective. We further discussed the brain structural changes over a set of prototypical brains as the explainable complement set from the morphological perspective. Finally, we investigated numerous scenarios from the perspective of their suitability for clinical applications by using several subsequent samples over the longitudinal 3D sMRIs, demonstrating AD progression over the years.

\begin{figure}
\centering
\begin{subfigure}{0.4\textwidth}
    \includegraphics[width=\textwidth]{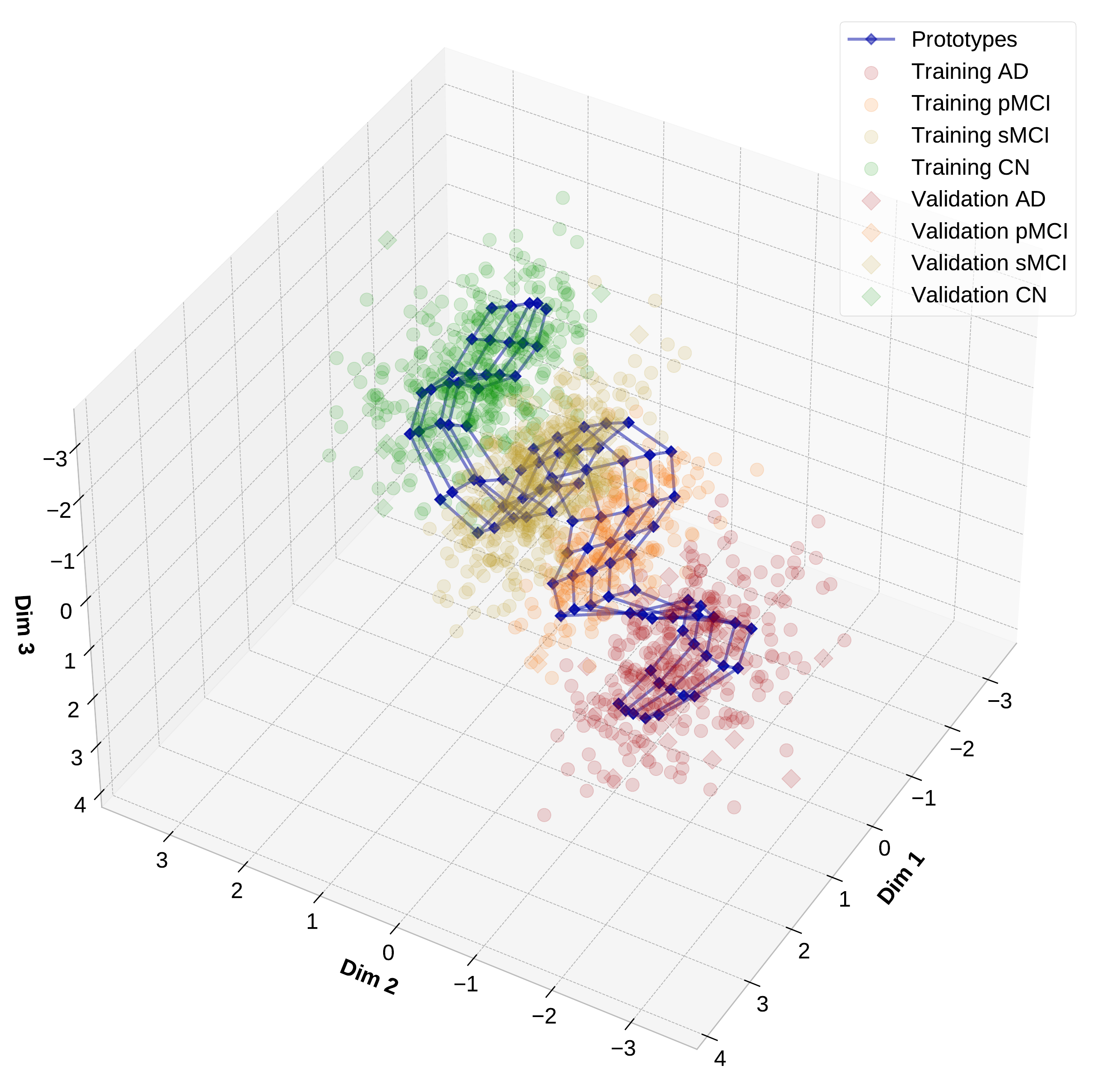}
    \caption{Established clinically-guided prototypes}
    \label{fig:adni_vaesom_decoded_space}
\end{subfigure}
\hfill
\begin{subfigure}{0.5\textwidth}
    \centering
    \begin{subfigure}{0.8\textwidth}
        \includegraphics[width=\textwidth]{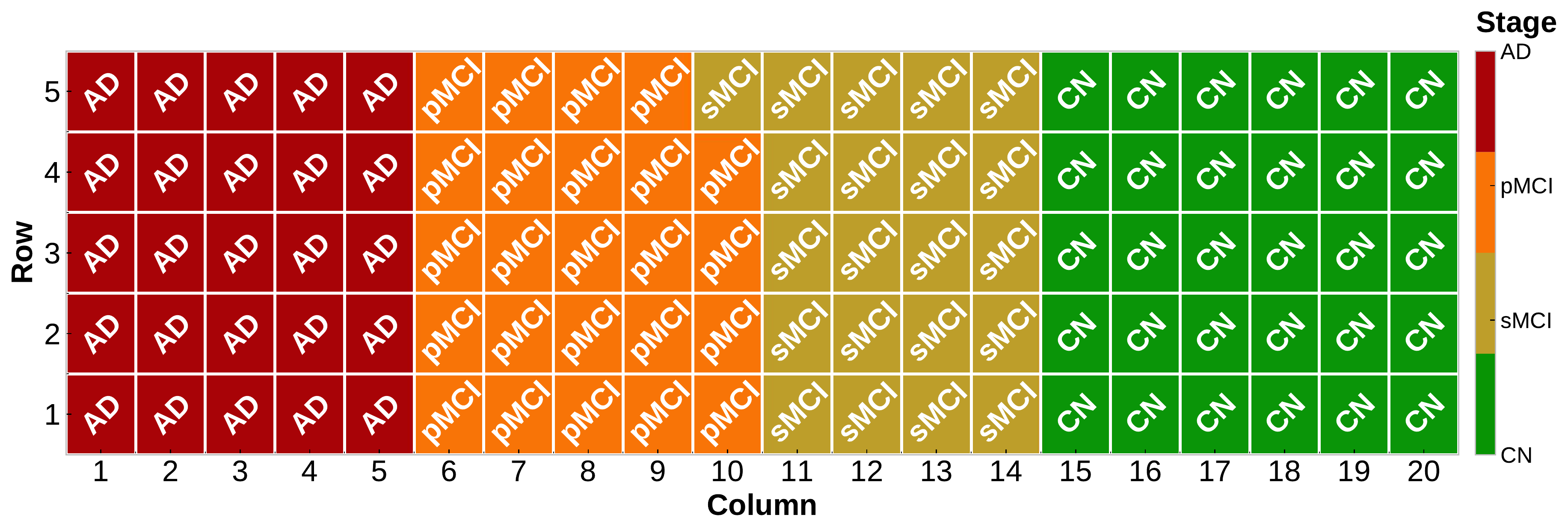}
        \vspace{-0.7cm}
        \caption{Prototypical clinical stage}
        \label{fig:adni_vaesom_decoded_stage}
    \end{subfigure}
    \hfill
    \begin{subfigure}{0.8\textwidth}
        \includegraphics[width=\textwidth]{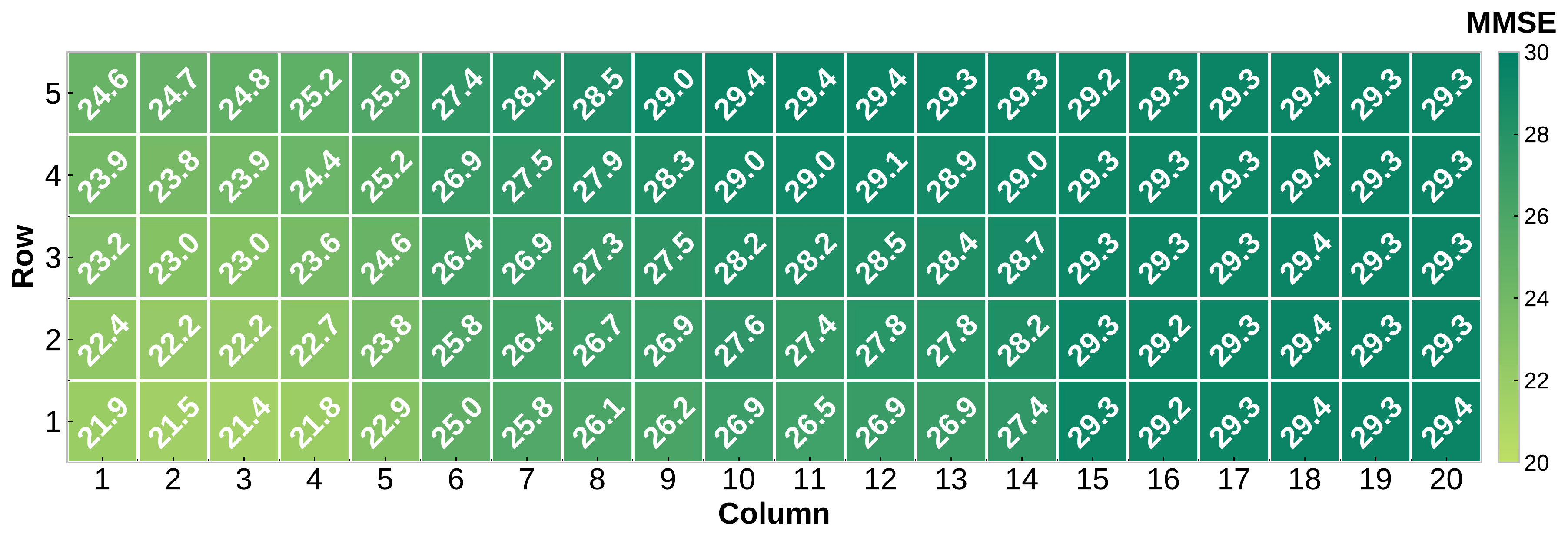}
        \vspace{-0.7cm}
        \caption{Prototypical cognitive score (MMSE)}
        \label{fig:adni_vaesom_decoded_mmse}
    \end{subfigure}
    \hfill
    \begin{subfigure}{0.8\textwidth}
        \includegraphics[width=\textwidth]{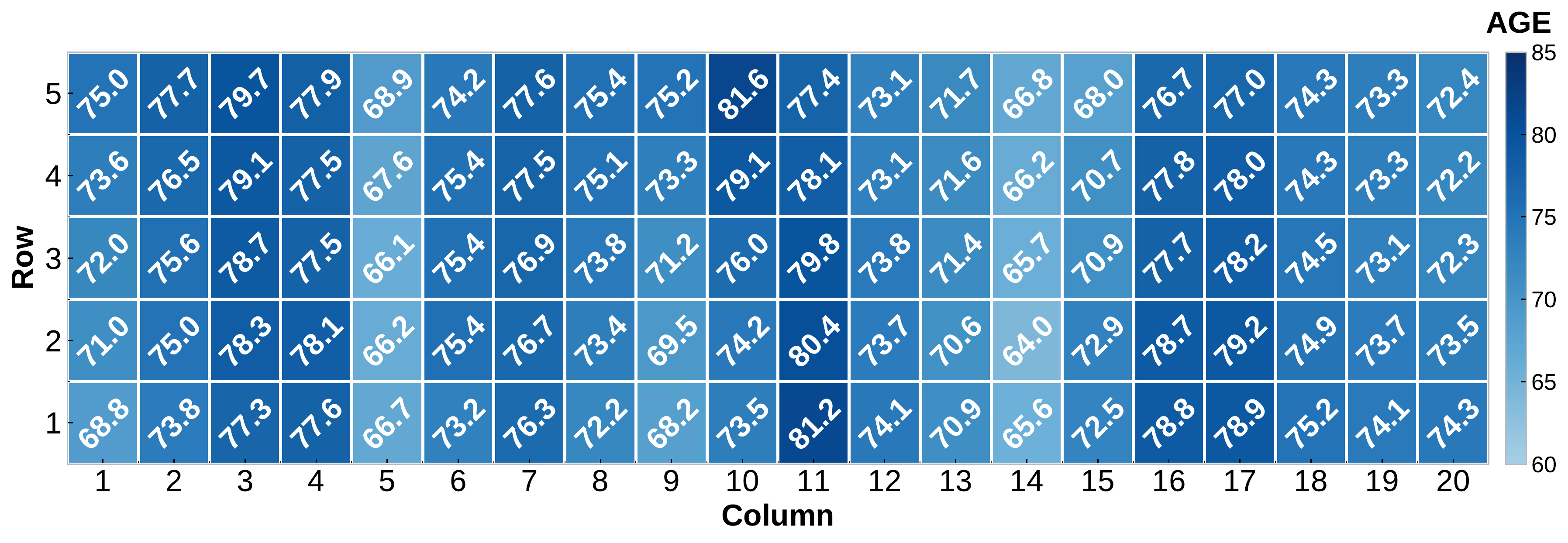}
        \vspace{-0.7cm}
        \caption{Prototypical age}
        \label{fig:adni_vaesom_decoded_age}
    \end{subfigure}
\end{subfigure}
\caption{Visualization of (a) latent clinical features and clinically-guided prototypes with respective decoded values providing a tuple of prototypical clinical states comprised of (b) clinical stage, (c) cognitive score, and (d) age.}
\label{fig:adni_vaesom_decoded}
\end{figure}

\begin{figure}
\centering
\begin{subfigure}{0.5\textwidth}
    \includegraphics[width=\textwidth]{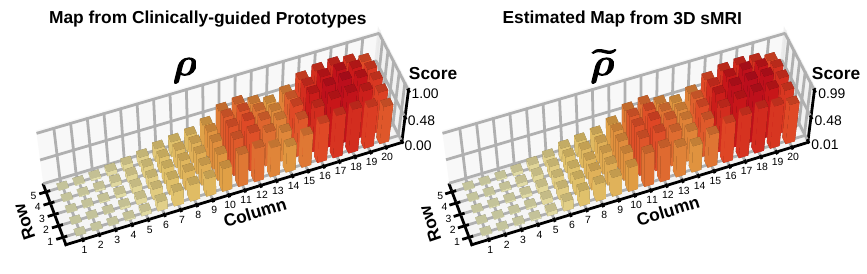}
    \caption{CN subject with MMSE of 30 and age of 75.70}
    \label{fig:estimated_AD_progression_map_cn}
\end{subfigure}
\hfill
\begin{subfigure}{0.5\textwidth}
    \includegraphics[width=\textwidth]{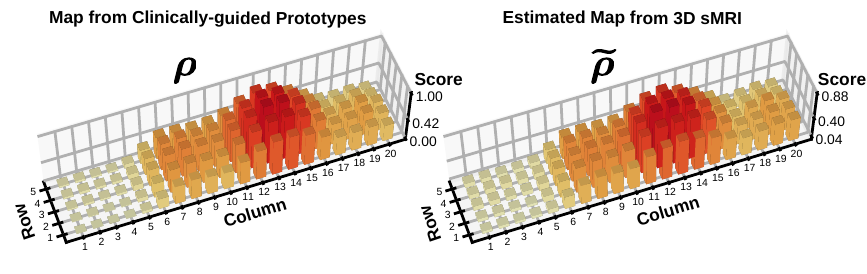}
    \caption{sMCI subject with MMSE of 29 and age of 74.10}
    \label{fig:estimated_AD_progression_map_smci}
\end{subfigure}
\begin{subfigure}{0.5\textwidth}
    \includegraphics[width=\textwidth]{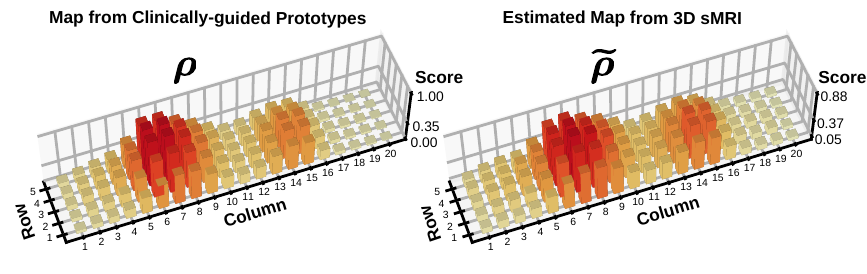}
    \caption{pMCI subject with MMSE of 27 and age of 84.20}
    \label{fig:estimated_AD_progression_map_pmci}
\end{subfigure}
\hfill
\begin{subfigure}{0.5\textwidth}
    \includegraphics[width=\textwidth]{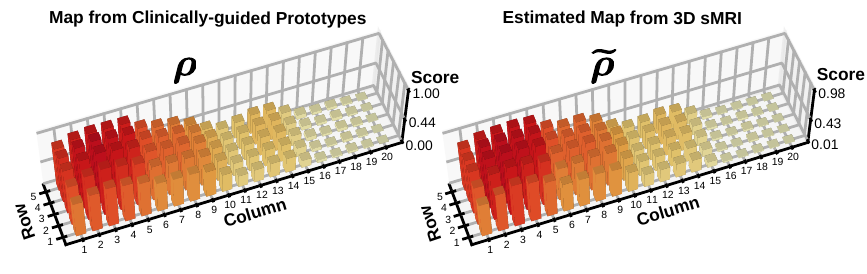}
    \caption{AD subject with MMSE of 23 and age of 80.70}
    \label{fig:estimated_AD_progression_map_ad}
\end{subfigure}
\caption{Illustration of the estimated AD likelihood map from a subject of (a) CN, (b) sMCI, (c) pMCI, and (d) AD. The heights and colors on each bar of $\vrho$ and $\widetilde{\vrho}$ represent the similarity or the resemblance of the clinical states to a set of prototypes, offering interpretability on existing clinical states of the given sMRI scan.}
\label{fig:estimated_AD_progression_map}
\end{figure}

\subsubsection{Clinically-guided Prototypes Establishment}
The experimental results on the clinically-guided prototypes establishment as the initial stream of our proposed \mbox{XADLiME} framework via ADPEN are depicted in Appendix D.3. Based on the joint visualization of the clusters of clinical features (we depicted the training and validation set) and a set of established prototypes, we observed that our proposed ADPEN effectively discovered the AD progression manifold and perceived the progression by thoroughly placing the prototypes across the AD spectrum (\ie, each color-coded according to their stages). As stated earlier, we reaffirmed that each of those clinically-guided prototypes plays a role as a representative in describing the subject's clinical states. Using the decoding network, we can reconstruct each prototype $\{\rvp_k^*\}_{k=1}^K$ to observe their clinical states. We represent those prototypical clinical states in a 2D topology scenario in Fig. \ref{fig:adni_vaesom_decoded}, with a tuple set of  clinical stage, cognitive score, and age. To some extent, the represented prototypical clinical states could be viewed as an enriched reference as they adequately captured a diverse range of samples reflecting possible AD progression pathways. For instance, the prototypical clinical stages reflected a set of stages for each clinical cluster spanned from CN towards AD across the AD spectrum (right to left in Fig. \ref{fig:adni_vaesom_decoded_stage}). Even though there was no actual direction of progression existed in those established prototypes, we could verify that the frameworks indeed complied with the AD spectrum given the nature of AD progression (\ie, a common AD spectrum progressed as CN$\rightarrow$sMCI$\rightarrow$pMCI$\rightarrow$AD with extremely rare instances of reverse conversion). Likewise, as the cognitive scores are, in fact, highly correlated with the clinical stages, we observed a declining pattern in such scores as it traverses towards AD (right to left in Fig. \ref{fig:adni_vaesom_decoded_mmse}). These results confirm that the established clinically-guided prototypes appropriately grasped the nature of AD traits. As for age (Fig. \ref{fig:adni_vaesom_decoded_age}), we noticed a quite faint age progression in each clinical stage. This is because AD could be manifested in a diverse range of ages without any prevailing pattern in each clinical stage over the AD spectrum. By further cross-checking these results with the summarized demographics (see Appendix A.1) regarding the age distribution, we argued that these prototypical ages succinctly reflected a comprehensive range of samples to some degree.

\begin{figure*}
\centering
\begin{subfigure}{0.49\textwidth}
    \includegraphics[width=\textwidth]{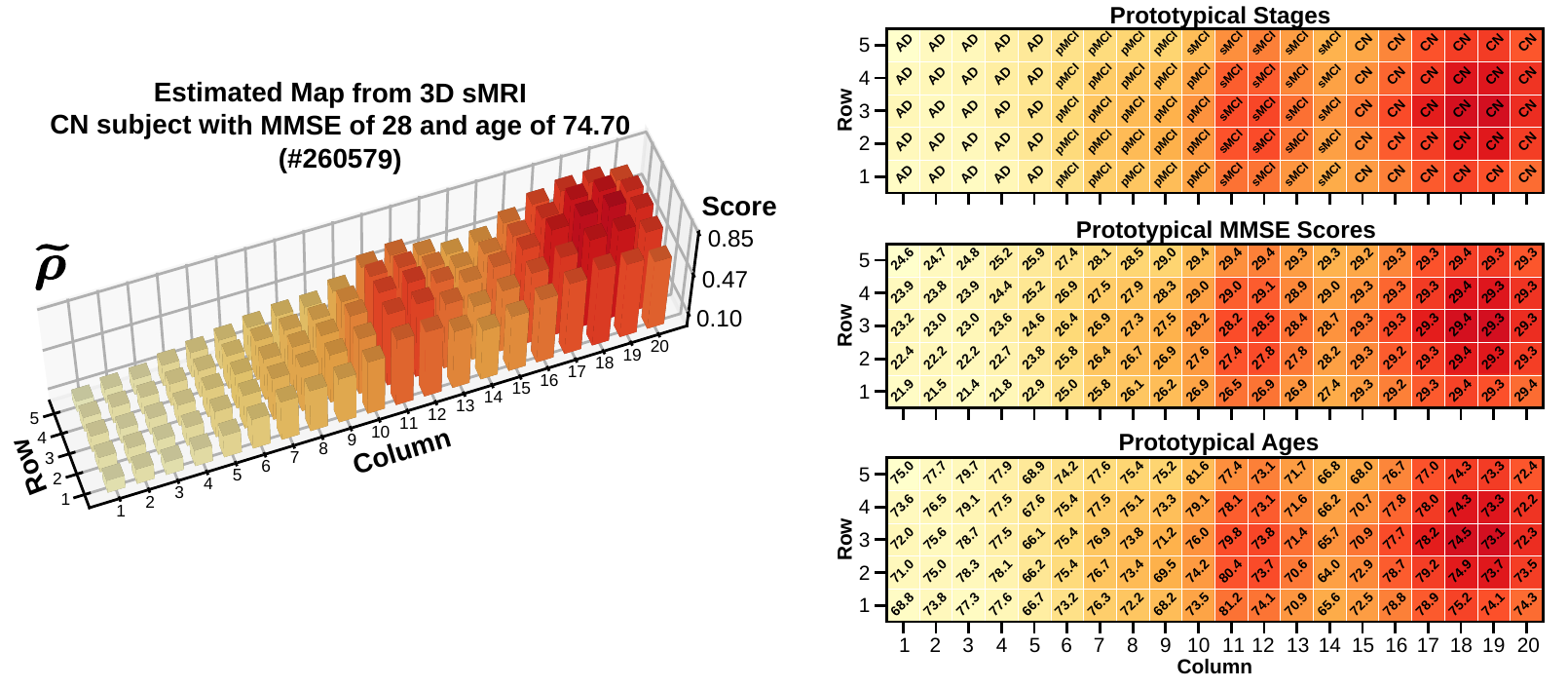}
    \caption{CN subject}
    \label{fig:adni_explainable_cn}
\end{subfigure}
\hfill
\begin{subfigure}{0.49\textwidth}
    \includegraphics[width=\textwidth]{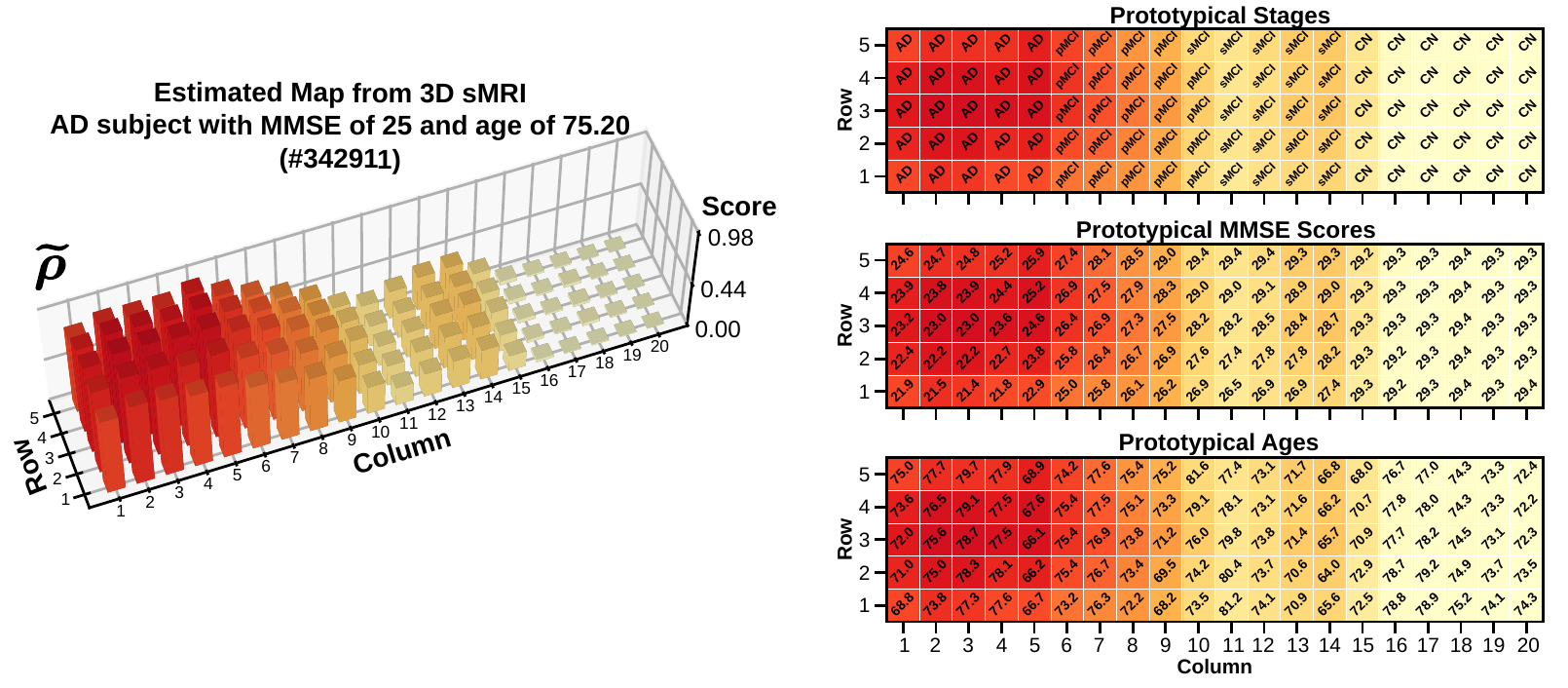}
    \caption{AD subject}
    \label{fig:adni_explainable_ad}
\end{subfigure}
\caption{Explainable AD likelihood maps of (a) CN and (b) AD subjects from the clinical perspectives.}
\label{fig:adni_explainable}
\end{figure*}

\begin{figure*}
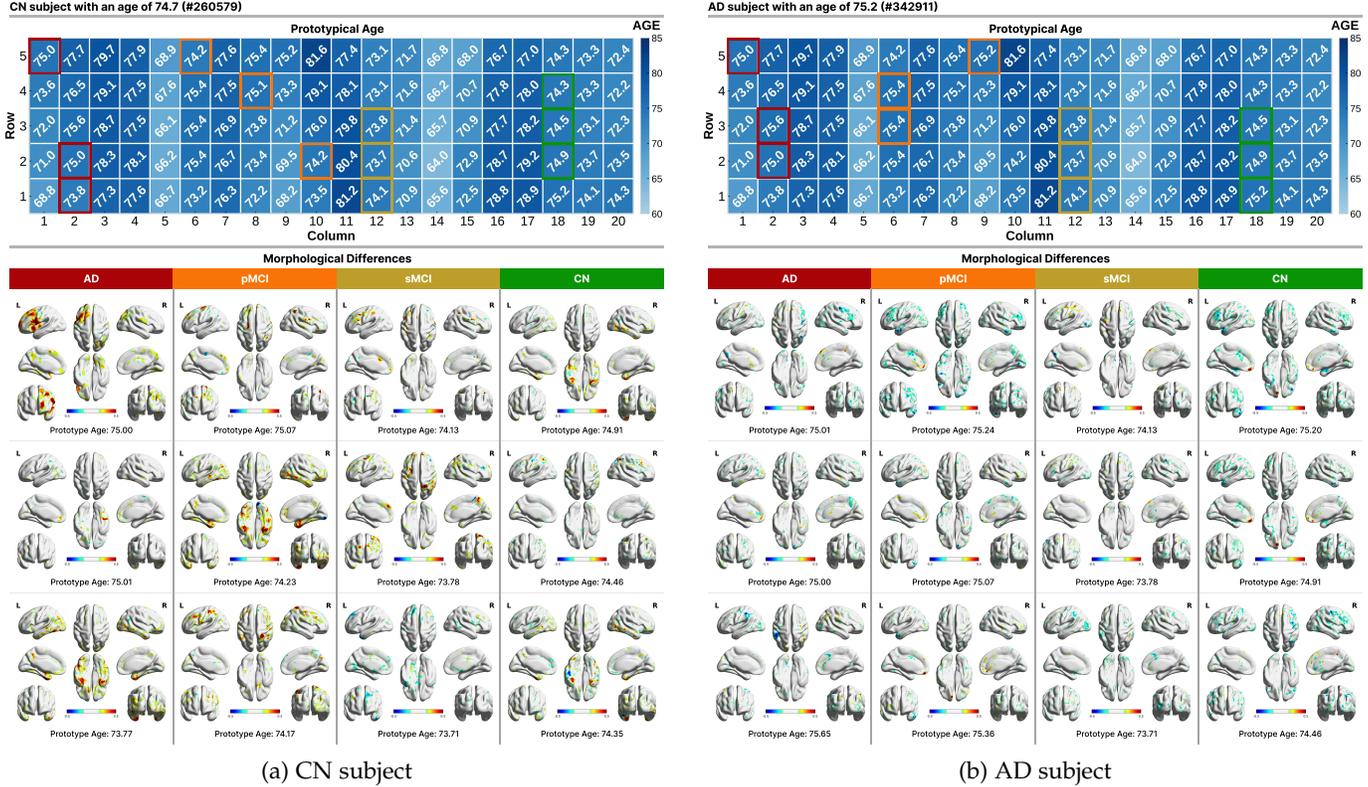

\centering
\begin{subfigure}{0.49\textwidth}
    \includegraphics[width=\textwidth]{figure/adni_morph_cn_260579.pdf}
    \caption{CN subject}
    \label{fig:adni_morph_cn}
\end{subfigure}
\hfill
\begin{subfigure}{0.49\textwidth}
    \includegraphics[width=\textwidth]{figure/adni_morph_ad_342911.pdf}
    \caption{AD subject}
    \label{fig:adni_morph_ad}
\end{subfigure}
\caption{Morphological changes visualization on cortical thickness by subtracting an unseen sample of a (a) CN subject with an age of 74.7 and (b) an AD subject with an age of 75.2 from the respective selected prototypical brains.}
\label{fig:adni_morph}
\end{figure*}

\subsubsection{Estimated Explainable AD Likelihood Map}
Having designed XADLiME as an explainable predictive framework, we visualized the AD likelihood maps with the 2D topology in Fig. \ref{fig:estimated_AD_progression_map}. To fully grasp the progression changes of such maps across the AD spectrum, we dispensed each sub-figure correspond to a sample in each stage (\ie CN, sMCI, pMCI, and AD). We utilized the 3D bar plot to visualize maps of $\vrho$ and $\widetilde{\vrho}$ as the pseudo-likelihood map measured from clinical information and the one estimated from 3D sMRI, accordingly. The heights and colors of each bar gave us the highlights of the underlying set of clinical states because it was built by measuring the similarity of latent clinical features to all established clinically-guided prototypes. As we specifically optimized the framework to exploit the ``pseudo" likelihood map $\vrho$ as our reliable assistance in estimating the respective map $\widetilde{\vrho}$ from the 3D sMRI, the conditions about underlying clinical states are still held in the estimated AD likelihood map as well. We confirmed the proficiency of XADLiME in transforming the 3D sMRI into the likelihood map of AD and provided explainability by means of a clinical-related ``heatmap" across the AD spectrum. However, such a heatmap alone does not comprehensively give us ease of interpretation from the clinical perspective. For accommodating this need, we substantially magnified its explainability by merging the estimated map with the prototypical clinical states. For this, we illustrated the final form of the explainable AD likelihood map on CN and AD subjects in Fig. \ref{fig:adni_explainable} as examples. In contrast with the ``raw" estimated AD likelihood map in Fig. \ref{fig:estimated_AD_progression_map}, as we jointly merged the comprehensive prototypical clinical states with the beneficial highlights provided by such a heatmap, we acquired an enhanced explainable AD likelihood map. Through this explainable map, we inferred plenty of information regarding the current clinical conditions of the subjects (\ie, clinical stage, MMSE, and age) and naturally determined the risk of potential progress towards the subsequent severe stage. For instance, we could observe in Fig. \ref{fig:adni_explainable_cn} that the map rendered slightly more attention in the sMCI domain, with most of the prototypical cognitive scores being highlighted around 28, which concisely describes the clinical conditions of the given subject. 

\subsubsection{Explainable Morphological Changes via Prototypical Brains}
In addition to performing the role of a guide for estimating AD likelihood maps, the established clinically-guided prototypes presumably held intrinsic progression-related facts over the brain sMRI scans morphologically. In particular, as each prototype enclosed a tuple of clinical information, we were able to retrieve the nearest respective values over the training samples and eventually obtain the corresponding sMRI scans. Henceforth, we refer to those sets of sMRI scans as \emph{prototypical brains}, which can be considered as the representative samples in an AD spectrum manifold, reflecting a group of subjects or a population. 
Through these prototypical brains, we further analyze the morphological changes map, which complements the clinical AD likelihood maps offered by our XADLiME framework. However, one of the prerequisites to properly simulate the morphological changes over such prototypical brains was that each brain region on it shall be registered or aligned into a common brain atlas or template in a non-linear manner. Thus, we additionally employed another pre-processing procedure over the ADNI dataset to determine and utilize the cortical thickness (CT) over the sMRI scans\footnote{Further details on the cortical thickness pre-processing procedures are provided in Appendix A.2.} \cite{tustison2014large}. We carefully selected the cortical thickness as it was considered a reliable AD-signature biomarker that has been extensively studied in identifying AD and its severity \cite{schwarz2016,racine2018}.

Furthermore, in order to adequately visualize such morphological changes, we devised an algorithm inspired by the nearest neighbor to select numerous prototypical brains according to the similarity over the actual age of unseen sample \wrt\ the prototypical ages. Note that given such an unseen sample of sMRI scan, the age of the respective subject will be most likely available as a common scenario in clinical applications. For this purpose, we initially picked the top-3 nearest corresponding sMRIs from training samples by measuring the distance of its clinical information to each unit of clinically-guided prototypes in the latent spaces (\ie, we then retrieved $K \times 3$ sMRIs). We priorly examined top-3 as a succinct number of samples to mitigate an undesirable outlier scenario. We further took an averaging operation over those top-3 retrieved sMRI samples, obtaining $K$ number of averaged scans. Subsequently, to accommodate the visual changes in mimicking the AD progression, we obtained three samples for each clinical stage that have the closest distances in terms of their age to the unseen test sample and further referred to such selected samples as the best representative prototypical brains. Once we obtained the corresponding prototypical brains, we subtracted the unseen sMRI sample from those sets of selected prototypical brains, treating the differences in cortical thickness as morphological changes. Afterward, we applied a necessary threshold operation to exhibit better focus in high-contrast values over the region of interests (ROIs). 

\begin{figure*}[!ht]
\centering
\begin{subfigure}{0.49\textwidth}
    \includegraphics[width=\textwidth]{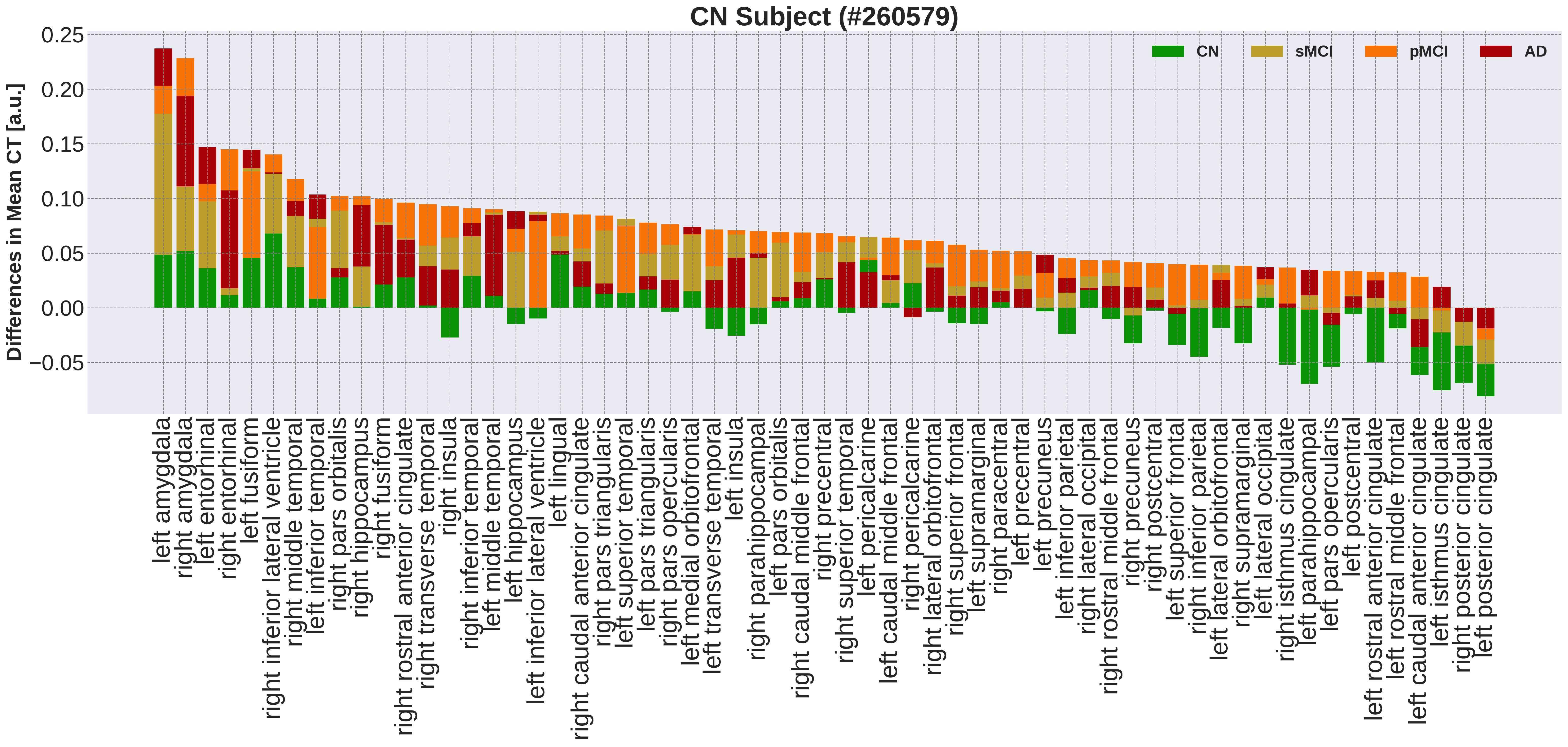}
    \vspace{-0.7cm}
    \caption{CN subject}
    \label{fig:roi_analysis_bar_cn}
\end{subfigure}
\hfill
\begin{subfigure}{0.49\textwidth}
    \includegraphics[width=\textwidth]{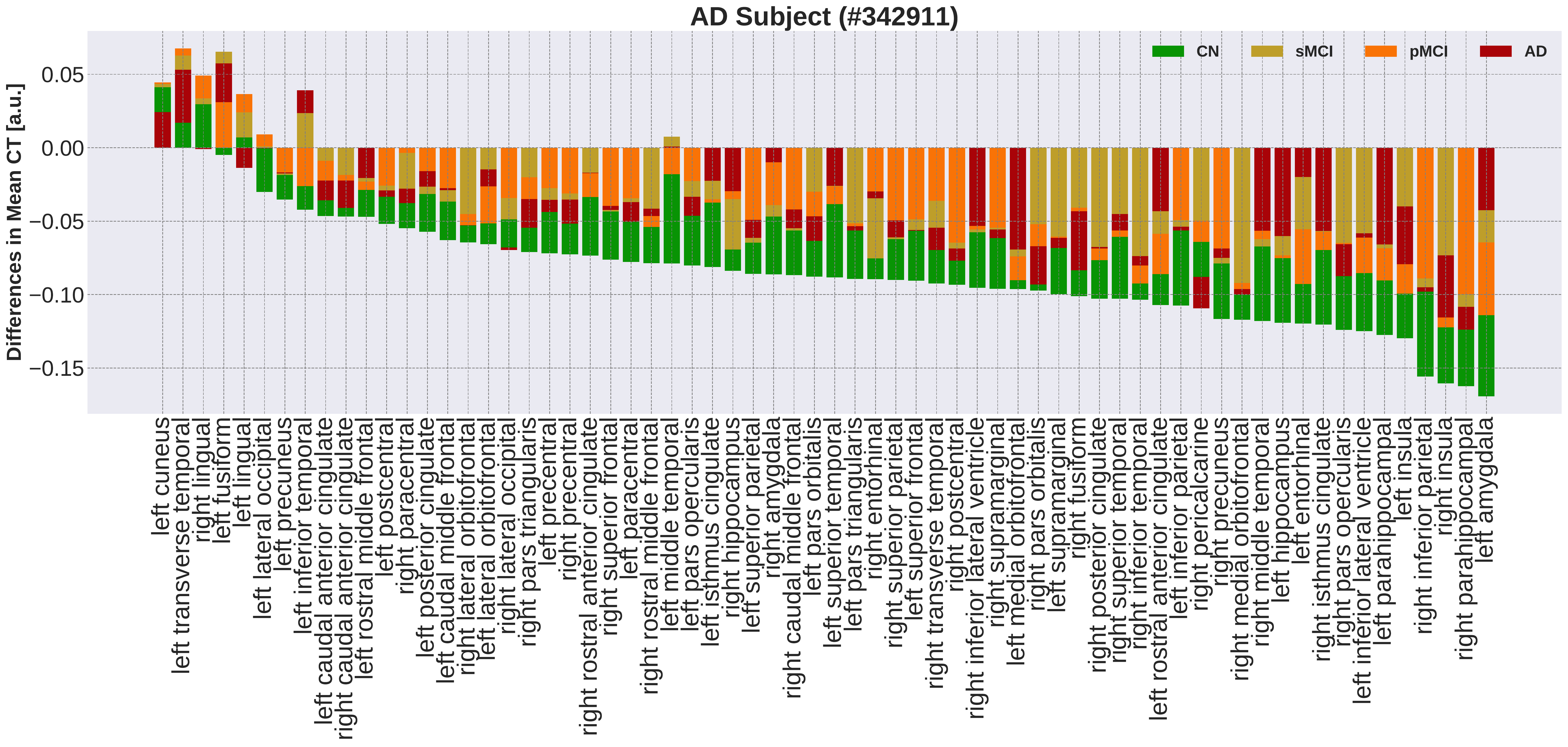}
    \vspace{-0.7cm}
    \caption{AD subject}
    \label{fig:roi_analysis_bar_ad}
\end{subfigure}
\caption{Visualization of the most highlighted ROIs based on the differences in CT of (a) CN and (b) AD samples subtracted from the prototypical brains. We ordered the obtained values to contrast the apparent pattern between two subjects.}
\label{fig:roi_analysis}
\end{figure*}

The discovered morphological changes are depicted in Fig. \ref{fig:adni_morph_cn} and Fig.  \ref{fig:adni_morph_ad} as a supplement to the explainable maps illustrated in Fig. \ref{fig:adni_explainable}. In the first scenario, the morphological changes over the prototypical brains were retrieved from an unseen sample of a CN subject with an age of 74.70, as depicted in Fig. \ref{fig:adni_morph_cn}. Here, we observed that the overall cortical thickness differences increase (red-highlighted) as we traverse from CN towards AD (right to left) in certain AD-related regions. In contrast, Fig. \ref{fig:adni_morph_ad} shows the visualization of morphological changes in an AD subject with an age of 75.2, which exhibits a tendency for the differences in cortical thickness to be lower (blue-highlighted) as we cross from the AD towards CN (left to right). Note that this morphological analysis relied upon a set of prototypical brains that were retrieved from multiple individuals; consequently, it comprised inevitable inter-subject variances. To further investigate and verify that the selected prototypical brains indeed captured the morphological changes through the AD-related ROIs of the brain, we present the averaged values (over the prototypical brains) of differences in cortical thickness group by each clinical stage in Fig. \ref{fig:roi_analysis} for such CN and AD subjects. Overall, the discovered contrasting patterns were considerably aligned with the prevailing facts regarding cortical thinning previously discovered by \cite{putcha2011,schwarz2016,racine2018,kulason2019cortical,yang2019study} in specific AD-signature ROIs. For instance, we observed relatively strong values (regardless of the sign) in cortical thickness differences over the amygdala, hippocampus, parahippocampal gyrus, temporal gyrus, superior frontal gyrus, precuneus, supramarginal gyrus, superior parietal gyrus, middle frontal gyrus, and so on. In real clinical applications, we could consider that the subject with the closest distances in terms of their clinical and morphological similarities to the selected prototypical brains may be treated using a similar approach (a thorough individual medical check-up shall still be administered carefully). Appendix D.4 provides results of MCI subjects.

\begin{figure*}[!ht]
\centering
\includegraphics[width=1.0\textwidth]{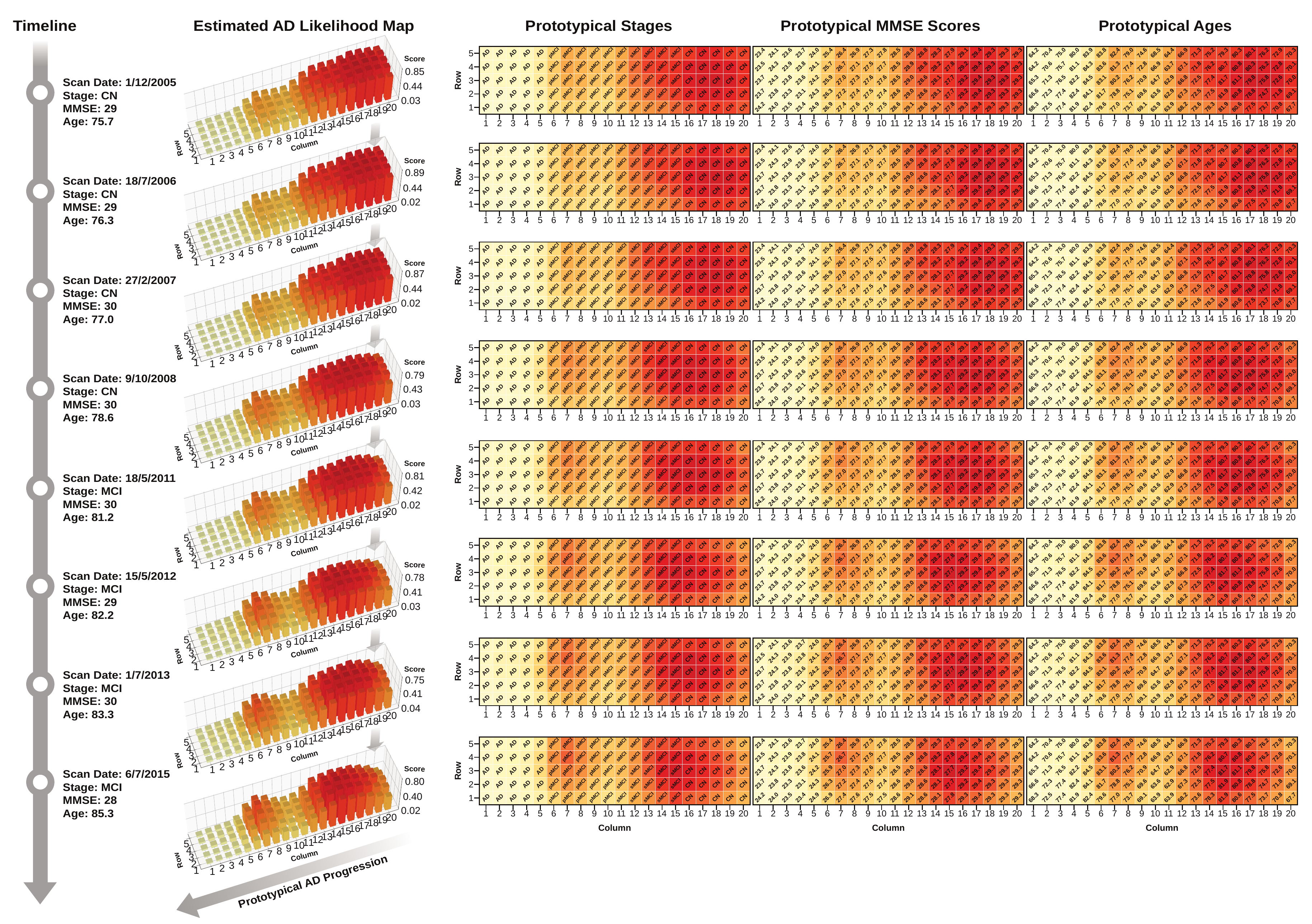}
\caption{Longitudinal scenario of a healthy subject progressing towards (stable) MCI throughout the follow-up years.}
\label{fig:adni_longitudinal_case_cn_mci}
\end{figure*}

\subsubsection{Potential Usage in Clinical Applications}
Rather than capturing the AD progression of subjects through their follow-up visits (\ie, a series of scans across consecutive years), the current framework was devised to perceive patterns of the prevailing AD progression in a population. Thus, we expedited an extensive feasibility investigation of our XADLiME in clinical applications for estimating the likelihood map from a single subject throughout the follow-up years by utilizing a series of longitudinal 3D sMRI scans and chronologically ordering them based on their acquisition date. Here, we treated such scans as the testing data without additional training over such longitudinal data. To provide a thorough analysis, we present the results of a longitudinal study on the prevalent case of a healthy subject who slowly progressed towards MCI.

We present a set of estimated AD likelihood maps in Fig. \ref{fig:adni_longitudinal_case_cn_mci} for the case of an elderly male subject with the age of 75.7 and MMSE of 29 on his first-visit acquisition date. This study was conducted between December 2005 to July 2015. He was diagnosed as a healthy subject during several follow-up visits (within three years from the initial scan). The estimated maps were able to capture such clinical facts by highlighting the CN prototypes with high scores. He was then diagnosed as an MCI subject for the remaining visits until his final visit in July 2015. Our framework could comprehend these disease trajectories as we observed the shifting of the heatmap towards the sMCI prototypical clinical stage throughout the subject's visits. We also observed noteworthy highlights, such as that there was an increase in the score over the pMCI prototypes once the subject was diagnosed as being an MCI subject, anticipating this likelihood as the typical case of AD progression. 

Through these results, we verified the potential usage of our proposed XADLiME framework in a scenario of real-world clinical application. Certainly, further exhaustive research shall be conducted to acquire a more robust model for handling numerous longitudinal cases. Additional longitudinal study cases are provided in Appendix D.5. 

\section{Conclusion}
We proposed XADLiME as a novel explainable predictive framework for modeling AD progression assisted by the clinically-guided prototypes. Such prototypes were established through our ADPEN by unifying VAE with a SOM module and the AD-spectrum-aware ordering loss. We also utilized a deep estimation model to transform 3D sMRI into an estimated AD likelihood map assisted by the similarities of the latent clinical features to the established clinically-guided prototypes. We then merged this estimated map with the decoded prototypical clinical states to substantially magnify its explainability in examining the clinical states manifested in a 3D sMRI. We analyzed the explainability from the morphological perspective by introducing a set of prototypical brains and thoroughly demonstrating the AD-related brain's morphological changes. Performance comparison on the downstream clinical tasks for classification as well as regression tasks over the ADNI cohort demonstrated the effectiveness of our proposed methods over the existing diagnostics methods while simultaneously providing ease of interpretation via the estimated explainable AD likelihood map. Finally, we presented several study cases exploiting longitudinal scans to investigate the potential usage in clinical applications, which turned out to perform well in simulating AD progression chronologically.

\section*{Acknowledgement}
This work was supported by Institute of Information \& communications Technology Planning \& Evaluation (IITP) grant funded by the Korea government (MSIT) No. 2022-0-00959 ((Part 2) Few-Shot Learning of Causal Inference in Vision and Language for Decision Making) and No. 2019-0-00079 (Artificial Intelligence Graduate School Program (Korea University)).

\bibliographystyle{IEEEtran}
\bibliography{main}


%





\ifCLASSOPTIONcaptionsoff
  \newpage
\fi



%

%

\begin{IEEEbiography}[{\includegraphics[width=1in,height=1.25in,clip,keepaspectratio]{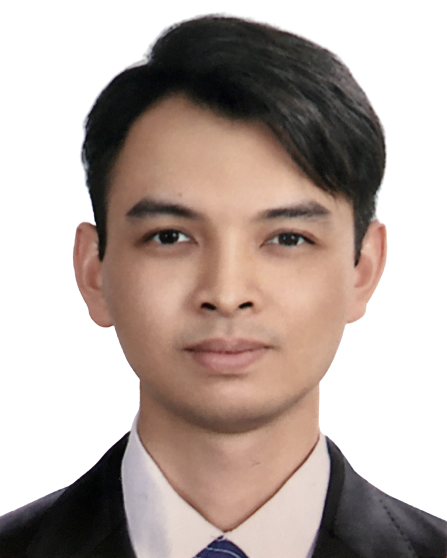}}]{Ahmad Wisnu Mulyadi}
received the bachelor’s degree in computer science education from the Indonesia University of Education, Bandung, Indonesia,  in 2010.  He is currently pursuing the Ph.D. degree with the Department of Brain and Cognitive Engineering, Korea University, Seoul, South Korea. 

His current research interests include machine, deep learning in healthcare, time series modeling, and biomedical image analysis.
\end{IEEEbiography}

\begin{IEEEbiography}[{\includegraphics[width=1in,height=1.25in,clip,keepaspectratio]{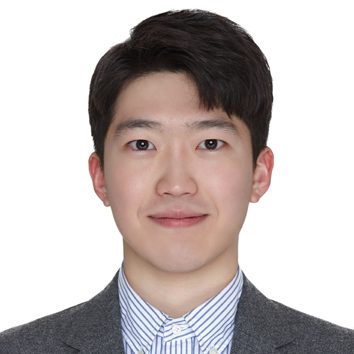}}]{Wonsik Jung}
	received the B.S. degree in Bio Medical Engineering from Konyang University, Daejeon, South Korea, in 2018. He is currently pursuing the Ph.D. degree with the Department of Brain and Cognitive Engineering, Korea University, Seoul, South Korea. 
    
    His current research interests include computer vision, time series modeling, and representation learning.
\end{IEEEbiography}

\begin{IEEEbiography}[{\includegraphics[width=1in,height=1.25in,clip,keepaspectratio]{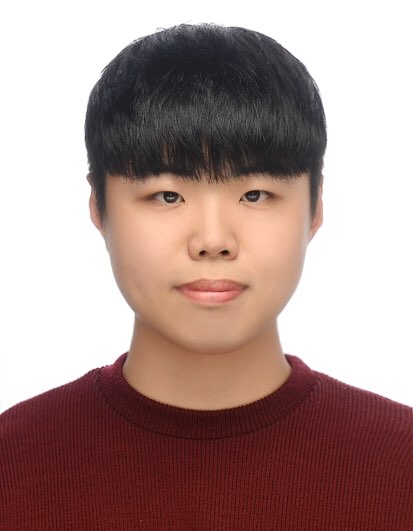}}]{Kwanseok Oh}
	received the B.S. degree in Electronic Control and Engineering from Hanbat National University, Daejeon, South Korea, in 2020. He is currently pursuing the M.S. degree with the Department of Artificial Intelligence, Korea University, Seoul, South Korea.
    
    His current research interests include computer vision, explainable AI, and machine/deep learning. 
\end{IEEEbiography}

\begin{IEEEbiography}[{\includegraphics[width=1in,height=1.25in,clip,keepaspectratio]{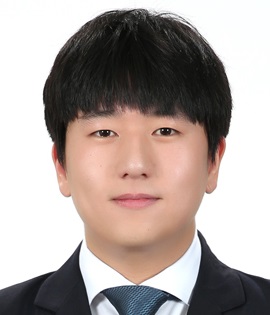}}]{Jee Seok Yoon}
	received the B.S. degree in Computer Science and Engineering from Korea University, Seoul, South Korea, in 2018. He is currently pursuing the Ph.D. degree with the Department of Brain and Cognitive Engineering, Korea University, Seoul, South Korea. 
    
    His current research interests include computer vision, meta learning, and representation learning. 
\end{IEEEbiography}

\begin{IEEEbiography}[{\includegraphics[width=1in,height=1.25in,clip,keepaspectratio]{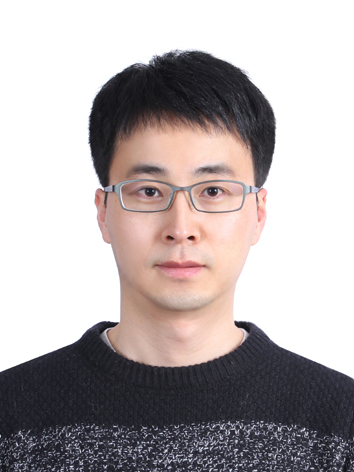}}]{Heung-Il Suk}
	received the Ph.D. degree in computer science and engineering from Korea University, Seoul, South Korea, in 2012.
	
	From 2012 to 2014, he was a Post-Doctoral Research Associate with the University of North Carolina at Chapel Hill, Chapel Hill, NC, USA. He is currently an Associate Professor with the Department of Artificial Intelligence and the Department of Brain and Cognitive Engineering, Korea University. 
    His research interests include machine/deep learning, explainable AI, biomedical data analysis, and brain-computer interface.

    Dr. Suk is serving as an Editorial Board Member for Electronics, Frontiers in Neuroscience, International Journal of Imaging Systems and Technology (IJIST), and a Program Committee or Reviewer for NeurIPS, ICML, ICLR, AAAI, IJCAI, MICCAI, AISTATS, \etc
\end{IEEEbiography}

\vfill









\section*{Supplementary material}

# XADLiME: eXplainable Alzheimer's Disease Likelihood Map Estimation via Clinically-guided Prototype Learning
## – *Supplementary Materials*

Ahmad Wisnu Mulyadi, Wonsik Jung, Kwanseok Oh, Jee Seok Yoon, and Heung-Il Suk, *Senior Member, IEEE*

---------------◆---------------

## APPENDIX A
## DATASET PREPROCESSING

### A.1 Data Preprocessing

Prior to the training, each 1.5/3T T1-weighted 3D sMRI scan of ADNI dataset [1] went through a series of preprocessing steps: neck removal process via *robustfov* provided in FMRIB Software Library (FSL) by Jenkinson *et al.* [2]; brain extraction using HDBet following Isensee *et al.* [3]; and linear registration via FMRIB Linear Registration Toolkit (FLIRT) provided by Jenkinson *et al.* [2]. We then applied the down-scaling by a factor of two ($2\times$), resulting in preprocessed 3D sMRI scan with a dimension of $96 \times 114 \times 96$. We employed the random cropping function over the 3D sMRI scans of the training set as an augmentation approach to mitigate the prevailing overfitting issue. Furthermore, we employed Quantile-normalization with $\{10\%, 90\%\}$, followed by a Gaussian normalization. As for the tuple of clinical information, we converted the clinical-stage label into a one-hot-encoding vector with $\mathbf{y} \in \mathbb{R}^4$, while normalized MMSE score and age of all sets via Gaussian normalization with mean and standard deviation measured from the training set. The demographic statistics of this cohort are summarized in Table 1.

### A.2 Cortical Thickness Preprocessing

In order to investigate the morphological changes using the cortical thickness of unseen samples by contrasting them with a set of selected prototypical brains, we additionally preprocessed the ADNI structural brain imaging (*i.e.*, 3D sMRI) dataset through the Advanced Normalization Tools

- A. W. Mulyadi, W. Jung, and J. S. Yoon are with the Department of Brain and Cognitive Engineering, Korea University, Seoul 02841, Republic of Korea (e-mail: {wisnumulyadi, ssikjeong1, wltjr1007}@korea.ac.kr).
- K. Oh is with the Department of Artificial Intelligence, Korea University, Seoul 02841, Republic of Korea (e-mail: ksohh@korea.ac.kr).
- H.-I. Suk is with the Department of Artificial Intelligence and the Department of Brain and Cognitive Engineering, Korea University, Seoul 02841, Republic of Korea and the corresponding author (e-mail: hisuk@korea.ac.kr).

TABLE 1: Demographic and clinical information summary of ADNI cohort.

| Aspect | CN | sMCI | pMCI | AD |
|---|---|---|---|---|
| #Subjects | 433 | 497 | 251 | 359 |
| Male/Female | 214/219 | 299/198 | 148/103 | 194/165 |
| Age (years) | 74.76±5.82 | 72.88±7.74 | 74.11±7.14 | 75.29±7.87 |
| Cognitive score | 29.07±1.12 | 27.83±1.76 | 26.79±1.74 | 23.21±2.05 |

(ANTs) toolbox[1]. Initially, we conducted a series of major preprocessing procedures over the sMRI scans with the following steps: i) N4 bias field correction; ii) brain extraction; iii) Atropos tissue segmentation for generating six tissues maps such as cortical gray matter, deep gray matter, white matter (WM), cerebrospinal fluid (CSF), brain stem, and cerebellum; iv) cortical thickness estimation, and v) non-linear registration to an MNI (Montreal Neurological Institute) template using the ANTs cortical thickness pipeline proposed by Tustison *et al.* [4]. Note that N4 was applied to i) and iii) during the process. Specifically, N4 produced an initial bias-corrected image for brain extraction. After extracting brain regions, N4 was repeatedly employed in the Atropos segmentation step to obtain the tissue probability map further.

## APPENDIX B
## NETWORK ARCHITECTURE

We describe a thorough details regarding the entire networks proposed in the XADLiME framework. For the sake of clarity, we denote several terms in relation to the layer's components: fully-connected layer (FC); convolutional layer (Conv) with its specific dimensionality (1D, 2D, or 3D), transposed one (T), kernel size ($\kappa$), stride ($\varsigma$), and padding ($\varrho$) also with a specific dimensionality (1D, 2D or 3D); a flattening operation (FLATTEN); and finally a series of activation functions: rectified linear unit (ReLU), exponential linear unit (ELU), sigmoid, and softmax. In

1. http://stnava.github.io/ANTs/

addition, to ensure the ease of information flow, we put the sequential number regarding the block, and the respective notation with the prime (′) indicates the intermediate notation that might not be introduced in the main manuscript.

### B.1 AD-topological-aware Prototypical Embedding Network (ADPEN)

A succinct overview of ADPEN is presented in Fig. 1. We built the ADPEN upon a VAE (*i.e.*, comprised of $\mathcal{E}_{\mathbf{c}}(.; \phi)$ (Table 2) and $\mathcal{D}_{\mathbf{c}}(.; \theta)$ (Table 4) as its encoding and decoding network, respectively) with two key components: (i) a lightweight trainable layer $\mathcal{O}(.; \psi)$ (Table 3) to embed and order the features using specified loss in the latent space and (ii) a SOM module to establish a set of prototypes with a predefined topological arrangement. Following are the detail of each component in the proposed ADPEN (excluding the SOM module as there is no specific architecture utilized).

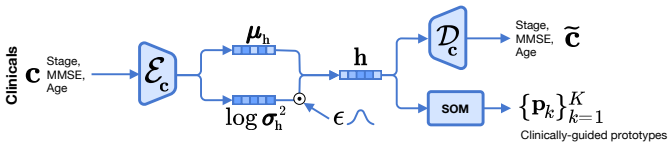

Fig. 1: Overview of our AD-spectrum-aware prototypical embedding network (ADPEN). It is composed of VAE augmented with an ordering loss along with a SOM module for clustering the latent clinical features to comprehend the AD spectrum while establishing a set of prototypes $\mathbf{P}$ in the latent space jointly.

TABLE 2: ADPEN encoding network $\mathcal{E}_{\mathbf{c}}(.; \phi)$.

| Block | Input | | Layer | Output | |
|---|---|---|---|---|---|
| | Symbol | $\mathbb{R}$ | | $\mathbb{R}$ | Symbol |
| | $\mathbf{y}$ | 4 | FC-ReLU | 8 | $\mathbf{y}'$ |
| Embedding | $\mathbf{c} = [\mathbf{y}' \circ s \circ a]$ | 10 | FC-ReLU | 10 | $\mathbf{h}'$ |
| #1 | $\mathbf{h}'$ | 10 | FC-ReLU | 16 | $\mathbf{h}'$ |
| #2 | $\mathbf{h}'$ | 16 | FC-ReLU | 8 | $\mathbf{h}'$ |
| | $\mathbf{h}'$ | 8 | FC | 3 | $\boldsymbol{\mu}_{\mathbf{h}}$ |
| #3 | $\mathbf{h}'$ | 8 | FC | 3 | $\log \boldsymbol{\sigma}_{\mathbf{h}}^2$ |

TABLE 3: ADPEN embedding layer $\mathcal{O}(.; \psi)$.

| Block | Input | | Layer | Output | |
|---|---|---|---|---|---|
| | Symbol | $\mathbb{R}$ | | $\mathbb{R}$ | Symbol |
| #1 | $\mathbf{h}$ | 3 | Conv1D[$\kappa(1 \times 3), \varsigma(1 \times 1)$] | 1 | $o$ |

TABLE 4: ADPEN decoding network $\mathcal{D}_{\mathbf{c}}(.; \theta)$.

| Block | Input | | Layer | Output | |
|---|---|---|---|---|---|
| | Symbol | $\mathbb{R}$ | | $\mathbb{R}$ | Symbol |
| #1 | $\mathbf{h}$ | 3 | FC-ReLU | 8 | $\mathbf{c}'$ |
| #2 | $\mathbf{c}'$ | 8 | FC-ReLU | 16 | $\mathbf{c}'$ |
| #3 | $\mathbf{c}'$ | 16 | FC-ReLU | 10 | $\mathbf{c}'$ |
| | $\mathbf{c}'$ | 10 | FC-softmax | 4 | $\widetilde{\mathbf{y}}$ |
| #4 | $\mathbf{c}'$ | 10 | FC | 1 | $\widetilde{s}$ |
| | $\mathbf{c}'$ | 10 | FC | 1 | $\widetilde{a}$ |

### B.2 Explainable AD Likelihood Map Estimatation (XADLiME)

Prior to fed the features into such diagnostic network, we utilized a lightweight autoencoder upon the estimated AD likelihood maps through $\mathcal{E}_{\boldsymbol{\rho}}$ (Table 5) and $\mathcal{D}_{\boldsymbol{\rho}}$ (Table 6). For estimating the AD likelihood map, we devised several key networks, including a feature extractor $\mathcal{E}_{\mathbf{X}}(.; \varphi)$ (Table 7) to extract the meaningful latent features given 3D sMRI; a feed-forward network $\mathcal{F}(.; \Omega)$ (Table 8) to estimate the likelihood map; and a task-dependent diagnostic network $\mathcal{C}(.; \Phi)$ (*e.g.*, classifier or regressor) (Table 9).

## APPENDIX C
## TRAINING HYPERPARAMETERS

For optimizing the ADPEN, we defined the $\lambda_1$ as the hyperparameter in controlling the learning ability of the SOM module as 0.01 in the pretraining phase. Afterward, we set $\lambda_1 = 1$ during the finetuning. We present the details of the $\Gamma_{\min}$ and $\Gamma_{\max}$ as the vicinity hyperparameters used in training the ADPEN for 1D, 2D, and 3D topology, respectively, in the following Table 10.

In addition, to optimize the XADLiME for conducting the downstream clinical tasks, we set the hyperparameters $\lambda_2$ and $\lambda_3$ in the following Table 11.

## APPENDIX D
## ADDITIONAL EXPERIMENTAL RESULTS

### D.1 Ablation Studies

We investigated several classification scenarios regarding the topological arrangement settings for ADPEN within the proposed XADLiME framework. We thoughtfully chose the clinical-stage classification for the downstream clinical task in this ablation study as it represents the imminent primary goal for the ADPM. The experimental results of our ablation studies on the XADLiME for all classification scenarios are reported in Table 12.

### D.2 MMSE and Age Prediction

In addition to the classification task, we have also performed MMSE and age predictions as the downstream clinical tasks. The experimental results for MMSE and age prediction are summarized in Table 13.

### D.3 Clinically-guided Prototypes Establishment

We provide the clinically-guided prototypes establishment for overall scenarios utilized in the ablation studies. In particular, we present the clinically-guided prototypes for 1D, 2D and 3D topology with $K = \{64, 100\}$ in Fig. 2 and 3, respectively.

### D.4 Morphological Changes via Prototypical Brains

In the following Fig. 4 and 5, we provide morphological changes of sMCI and pMCI subjects in contrast with prototypical brains, respectively. Overall, the sMCI case exhibited relatively high values (red) towards AD while highlighted lower values (blue) over CN samples. These prevailing patterns were also retained for pMCI samples with significantly higher cortical differences over CN samples.

TABLE 5: Likelihood map encoding network $\mathcal{E}_{\rho}$. We further breakdown the total number of prototypes $K$ into $K_1, K_2$, and $K_3$ depending on the predefined topological arrangement (aforementioned in the implementation settings and ablation studies).

| Block | Input | | Layer | Output | |
|---|---|---|---|---|---|
| | Symbol | $\mathbb{R}$ | | $\mathbb{R}$ | Symbol |
| #1 | $\widetilde{\rho}$ | $K_1 \times K_2 \times K_3$ | Conv2D[$\kappa(K_2 \times 5)$, $\varsigma(1 \times 1)$]-BN2D | $(K_1 * 64) \times 1 \times (K_3 - 4)$ | $\mathcal{E}_{\rho}(\widetilde{\rho})$ |

TABLE 6: Likelihood map decoding network $\mathcal{D}_{\rho}$.

| Block | Input | | Layer | Output | |
|---|---|---|---|---|---|
| | Symbol | $\mathbb{R}$ | | $\mathbb{R}$ | Symbol |
| #1 | $\mathcal{E}_{\rho}(\widetilde{\rho})$ | $(K_1 * 64) \times 1 \times (K_3 - 4)$ | ConvT2D[$\kappa(K_2 \times 5)$, $\varsigma(1 \times 1)$]-BN2D-ReLU | $(K_1 * 64) \times K_2 \times K_3$ | $\widetilde{\rho}'$ |
| #2 | $\widetilde{\rho}'$ | $(K_1 * 64) \times K_2 \times K_3$ | Conv2D[$\kappa(1 \times 1)$, $\varsigma(1 \times 1)$]-sigmoid | $K_1 \times K_2 \times K_3$ | $\widetilde{\rho}$ |

TABLE 7: Feature extractor network $\mathcal{E}_{\mathbf{X}}(.; \varphi)$.

| Block | Input | | Layer | Output | |
|---|---|---|---|---|---|
| | Symbol | $\mathbb{R}$ | | $\mathbb{R}$ | Symbol |
| #1 | $\mathbf{X}$ | $1 \times 96 \times 114 \times 96$ | Conv3D[$\kappa(3 \times 3 \times 3)$, $\varsigma(2 \times 2 \times 2)$]-BN3D-ELU | $32 \times 47 \times 56 \times 47$ | $\mathbf{z}'$ |
| | $\mathbf{z}'$ | $32 \times 47 \times 56 \times 47$ | Conv3D[$\kappa(3 \times 3 \times 3)$, $\varsigma(1 \times 1 \times 1)$, $\varrho(1 \times 1 \times 1)$]-BN3D-ELU | $32 \times 47 \times 56 \times 47$ | $\mathbf{z}'$ |
| #2 | $\mathbf{z}'$ | $32 \times 47 \times 56 \times 47$ | Conv3D[$\kappa(3 \times 3 \times 3)$, $\varsigma(2 \times 2 \times 2)$]-BN3D-ELU | $64 \times 23 \times 27 \times 23$ | $\mathbf{z}'$ |
| | $\mathbf{z}'$ | $64 \times 23 \times 27 \times 23$ | Conv3D[$\kappa(3 \times 3 \times 3)$, $\varsigma(1 \times 1 \times 1)$, $\varrho(1 \times 1 \times 1)$]-BN3D-ELU | $64 \times 23 \times 27 \times 23$ | $\mathbf{z}'$ |
| #3 | $\mathbf{z}'$ | $64 \times 23 \times 27 \times 23$ | Conv3D[$\kappa(3 \times 3 \times 3)$, $\varsigma(2 \times 2 \times 2)$]-BN3D-ELU | $128 \times 11 \times 13 \times 11$ | $\mathbf{z}'$ |
| | $\mathbf{z}'$ | $128 \times 11 \times 13 \times 11$ | Conv3D[$\kappa(3 \times 3 \times 3)$, $\varsigma(1 \times 1 \times 1)$, $\varrho(1 \times 1 \times 1)$]-BN3D-ELU | $128 \times 11 \times 13 \times 11$ | $\mathbf{z}'$ |
| #4 | $\mathbf{z}'$ | $128 \times 11 \times 13 \times 11$ | Conv3D[$\kappa(3 \times 3 \times 3)$, $\varsigma(2 \times 2 \times 2)$]-BN3D-ELU | $256 \times 5 \times 6 \times 5$ | $\mathbf{z}'$ |
| | $\mathbf{z}'$ | $256 \times 5 \times 6 \times 5$ | Conv3D[$\kappa(3 \times 3 \times 3)$, $\varsigma(1 \times 1 \times 1)$, $\varrho(1 \times 1 \times 1)$]-BN3D-ELU | $256 \times 5 \times 6 \times 5$ | $\mathbf{z}'$ |
| #5 | $\mathbf{z}'$ | $256 \times 5 \times 6 \times 5$ | Conv3D[$\kappa(5 \times 6 \times 5)$, $\varsigma(1 \times 1 \times 1)$]-BN3D-ELU | $512 \times 1 \times 1 \times 1$ | $\mathbf{z}'$ |
| #6 | $\mathbf{z}'$ | $512 \times 1 \times 1 \times 1$ | FLATTEN | $512$ | $\mathbf{z}$ |

TABLE 8: Estimating network $\mathcal{F}(.; \Omega)$.

| Block | Input | | Layer | Output | |
|---|---|---|---|---|---|
| | Symbol | $\mathbb{R}$ | | $\mathbb{R}$ | Symbol |
| #1 | $\mathbf{z}$ | $512$ | FC-BN1D-ELU | $256$ | $\widetilde{\rho}'$ |
| #2 | $\widetilde{\rho}'$ | $256$ | FC-sigmoid | $K$ | $\widetilde{\rho}$ |

TABLE 9: Task-dependant diagnostic network $\mathcal{C}(.; \Phi)$. We set the activation function as softmax for classification and sigmoid for regression tasks, accordingly. The symbol * denotes that the input dimension varies depending on $K$ with its topological arrangement.

| Block | Input | | Layer | Output | |
|---|---|---|---|---|---|
| | Symbol | $\mathbb{R}$ | | $\mathbb{R}$ | Symbol |
| #1 | $\mathcal{E}_{\rho}(\widetilde{\rho})$ | * | FC-BN1D-ELU | $64$ | $\mathcal{E}_{\rho}(\widetilde{\rho})'$ |
| #2 (Binary classification) | $\mathcal{E}_{\rho}(\widetilde{\rho})'$ | $64$ | FC-softmax | $2$ | $\widetilde{\mathbf{y}}$ |
| #2 (Multiclass classification) | $\mathcal{E}_{\rho}(\widetilde{\rho})'$ | $64$ | FC-softmax | $3$ | $\widetilde{\mathbf{y}}$ |
| #2 (MMSE Prediction) | $\mathcal{E}_{\rho}(\widetilde{\rho})'$ | $64$ | FC-sigmoid | $1$ | $\widetilde{s}$ |
| #2 (Age Prediction) | $\mathcal{E}_{\rho}(\widetilde{\rho})'$ | $64$ | FC-sigmoid | $1$ | $\widetilde{a}$ |

## D.5 Longitudinal Study Cases

We report additional longitudinal study cases of a three healthy subjects who progressed towards AD during their clinical visits in Fig. 6.

In addition, in Fig. 7, we depict the estimated AD likelihood maps from 3D sMRI scans belonging to a male MCI subject with the age of 60.7 and MMSE score of 29 on his first visit in February 2012. At the initial time point, this map indicated the highlights mostly on sMCI with a hazy portion on the CN region. Throughout his visits, he was consistently diagnosed as an MCI subject within the following two years with a fluctuating MMSE score around 29–30. In February 2016, he was still diagnosed as an MCI subject at the final visit point. Our estimated map captured these stable MCI traits by steadily highlighting the sMCI in the prototypical

TABLE 10: Vicinity hyperparameters in training the proposed ADPEN.

| K | Topology | Pretraining | | Finetuning | |
|---|---|---|---|---|---|
| | | $\Gamma_{max}$ | $\Gamma_{min}$ | $\Gamma_{max}$ | $\Gamma_{min}$ |
| 64 | $1D : 64$ | 64.0 | 2.0 | 2.0 | 0.8 |
| 64 | $2D : 4 \times 16$ | 16.0 | 2.0 | 2.0 | 0.8 |
| 64 | $3D : 2 \times 2 \times 16$ | 16.0 | 2.0 | 2.0 | 0.8 |
| 100 | $1D : 100$ | 100.0 | 10.0 | 10.0 | 1.0 |
| 100 | $2D : 5 \times 20$ | 20.0 | 2.0 | 2.0 | 0.8 |
| 100 | $3D : 2 \times 2 \times 25$ | 25.0 | 2.0 | 2.0 | 0.8 |

TABLE 11: Hyperparameters to weigh the $\mathcal{L}_{Cons}$ and $\mathcal{L}_{Task}$.

| Clinical Tasks | $\lambda_2$ | $\lambda_3$ |
|---|---|---|
| Clinical Stage Classification | 1 | 1 |
| MMSE Prediction | 10 | 10 |
| Age Prediction | 1 | 1000 |

TABLE 12: Experimental results of ablation studies on ADNI dataset.

| Scenario | $K$ | Topology | AUC | Bal. ACC | F1-Score |
|---|---|---|---|---|---|
| CN/AD | 64 | $1D : 64$ | 0.9314±0.0410 | 0.8722±0.0458 | 0.8589±0.0509 |
| | 64 | $2D : 4 \times 16$ | 0.9423±0.0338 | 0.9067±0.0229 | 0.8972±0.0254 |
| | 64 | $3D : 2 \times 2 \times 16$ | 0.9308±0.0349 | 0.8903±0.0500 | 0.8771±0.0577 |
| | 100 | $1D : 100$ | 0.9395±0.0262 | 0.9041±0.0311 | 0.8936±0.0352 |
| | 100 | $2D : 5 \times 20$ | **0.9492±0.0330** | **0.9183±0.0320** | **0.9102±0.0356** |
| | 100 | $3D : 2 \times 2 \times 25$ | 0.9405±0.0301 | 0.8932±0.0415 | 0.8822±0.0463 |
| CN/MCI | 64 | $1D : 64$ | 0.7294±0.0337 | 0.6686±0.0416 | 0.7789±0.0287 |
| | 64 | $2D : 4 \times 16$ | 0.7457±0.0105 | 0.6767±0.0254 | **0.7886±0.0102** |
| | 64 | $3D : 2 \times 2 \times 16$ | 0.7461±0.0452 | 0.6466±0.0320 | 0.7825±0.0241 |
| | 100 | $1D : 100$ | 0.7405±0.0276 | 0.6725±0.0303 | 0.7619±0.0422 |
| | 100 | $2D : 5 \times 20$ | **0.7517±0.0355** | **0.6905±0.0439** | 0.7747±0.0458 |
| | 100 | $3D : 2 \times 2 \times 25$ | 0.7461±0.0505 | 0.6556±0.0322 | 0.7712±0.0440 |
| MCI/AD | 64 | $1D : 64$ | 0.7671±0.0578 | 0.6717±0.0657 | 0.5346±0.1133 |
| | 64 | $2D : 4 \times 16$ | 0.7708±0.0629 | 0.6801±0.0590 | 0.5517±0.0968 |
| | 64 | $3D : 2 \times 2 \times 16$ | 0.7699±0.0867 | 0.7088±0.0684 | 0.5954±0.1016 |
| | 100 | $1D : 100$ | 0.7583±0.0697 | 0.6724±0.0512 | 0.5381±0.0889 |
| | 100 | $2D : 5 \times 20$ | **0.7758±0.0584** | **0.7221±0.0530** | **0.6173±0.0742** |
| | 100 | $3D : 2 \times 2 \times 25$ | 0.7741±0.0661 | 0.7131±0.0664 | 0.6013±0.0973 |
| sMCI/pMCI | 64 | $1D : 64$ | 0.7478±0.0342 | 0.6664±0.0453 | 0.5405±0.0780 |
| | 64 | $2D : 4 \times 16$ | 0.7507±0.0319 | 0.6664±0.0144 | 0.5413±0.0362 |
| | 64 | $3D : 2 \times 2 \times 16$ | 0.7358±0.0497 | 0.6767±0.0306 | 0.5547±0.0447 |
| | 100 | $1D : 100$ | 0.7339±0.0217 | 0.6625±0.0364 | 0.5328±0.0617 |
| | 100 | $2D : 5 \times 20$ | **0.7567±0.0582** | **0.6785±0.0445** | **0.5576±0.0732** |
| | 100 | $3D : 2 \times 2 \times 25$ | 0.7322±0.0478 | **0.6785±0.0559** | 0.5455±0.1115 |
| CN/MCI/AD | 64 | $1D : 64$ | 0.7425±0.0430 | 0.6068±0.0366 | 0.5827±0.0462 |
| | 64 | $2D : 4 \times 16$ | 0.7526±0.0381 | 0.5900±0.0358 | 0.5665±0.0455 |
| | 64 | $3D : 2 \times 2 \times 16$ | 0.7471±0.0500 | 0.5875±0.0423 | 0.5657±0.0491 |
| | 100 | $1D : 100$ | 0.7542±0.0282 | 0.5923±0.0312 | 0.5706±0.0401 |
| | 100 | $2D : 5 \times 20$ | **0.7652±0.0386** | **0.6316±0.0519** | **0.6103±0.0573** |
| | 100 | $3D : 2 \times 2 \times 25$ | 0.7515±0.0319 | 0.6042±0.0215 | 0.5836±0.0252 |

TABLE 13: MMSE and age prediction performances on the ADNI dataset. The arrow symbol ↓ indicates the lower, the better, and vice versa in the case of ↑.

| Models | MMSE | | Age | |
|---|---|---|---|---|
| | RMSE (↓) | $R^2$(↑) | RMSE (↓) | $R^2$(↑) |
| ResNet-18 | 2.3122±0.1098 | 0.2921±0.0776 | 3.9386±0.3148 | 0.3762±0.1344 |
| Jin *et al.* | 2.3122±0.1584 | 0.2895±0.1132 | 3.9555±0.2563 | 0.3763±0.1003 |
| Korolev *et al.* | 2.2960±0.1988 | 0.3001±0.1200 | 4.2214±0.2684 | 0.2796±0.1738 |
| Liu *et al.* | 2.2310±0.2027 | 0.3380±0.1234 | 3.7927±0.5664 | 0.4099±0.1961 |
| **XADLiME** | **2.1892±0.1648** | **0.3657±0.0869** | **3.7704±0.2998** | **0.4246±0.1447** |

clinical stage across the years. Furthermore, we observed that the coverage on sMCI is shrinking over time, suggesting the growing fidelity in estimating that the given subject was

indeed a stable MCI subject.

We also illustrate the longitudinal case of subject progression towards AD from MCI in Fig. 8. Here, a male

subject with the age of 82.3 and MMSE of 28 was inspected as an MCI subject in his initial sMRI scanning date in July 2012. Our XADLiME estimated the likelihood map, which strongly highlights the pMCI clusters with a quite faint emphasis on a small portion of sMCI and AD, indicating the cautiousness in estimating the accurate clinical stage. Fast forward to July 2013 as the second to the last time point, the estimated map indicates a more potent concentration in the AD region but still highlighted mainly over the pMCI. To this end, the aforementioned faint highlights in sMCI were greatly diminished. These patterns suggested the anticipation of switching stages over the AD spectrum, which was adequately perceived by our XADLiME. Ultimately, two years later, in July 2015, his ultimate diagnosis was AD, with a significant decline in MMSE (*i.e.*, from 29 to 26). This fact was also successfully captured in the respective estimated map with more intense highlights in AD regions than at the last scanning point.

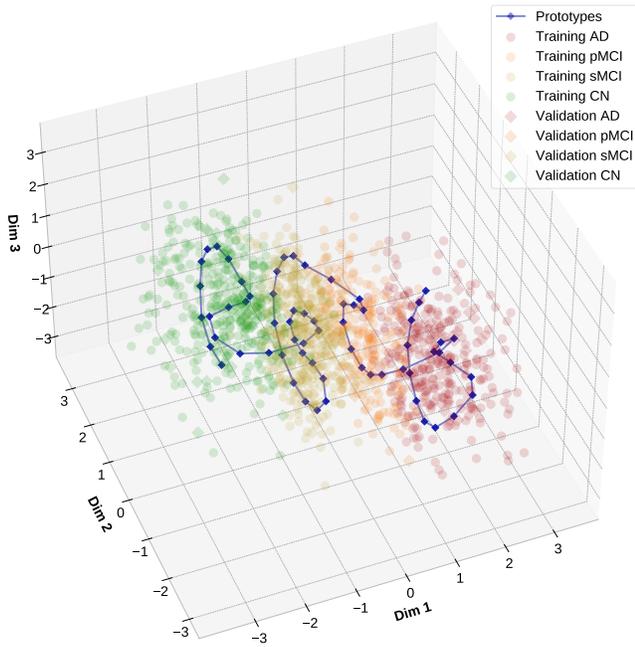

(a) 1D topology, $K = 64$

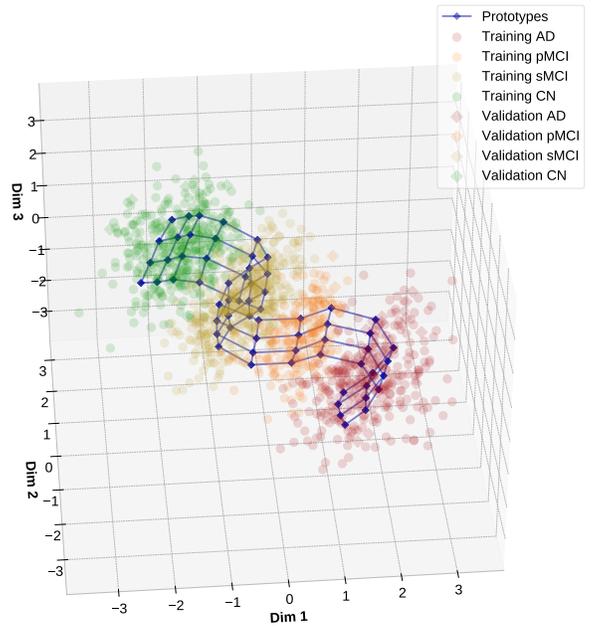

(b) 2D topology, $K = 64$

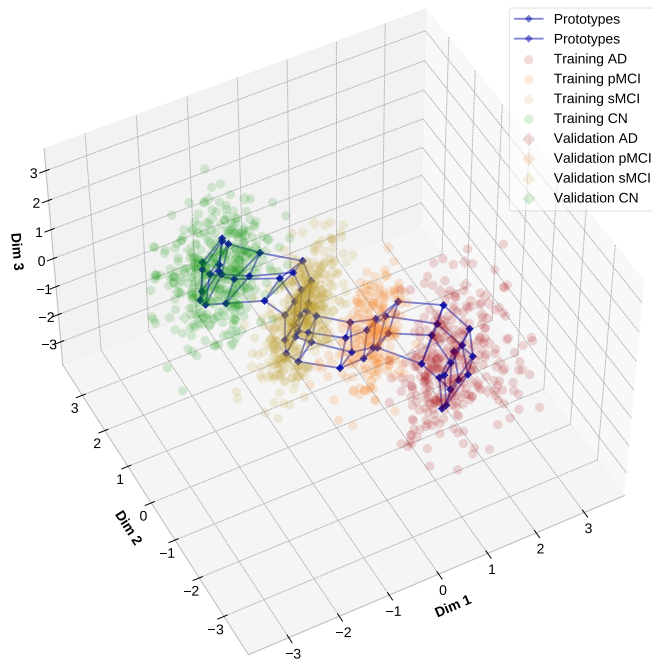

(c) 3D topology, $K = 64$

Fig. 2: Visualization of latent clinical features along with learned clinically-guided prototypes with (a) 1D, (b) 2D and (c) 3D topological arrangements using $K = 64$.

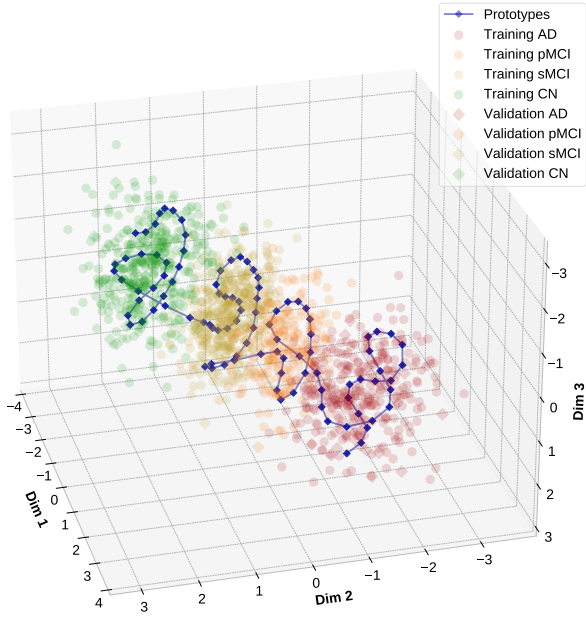

(a) 1D topology, $K = 100$

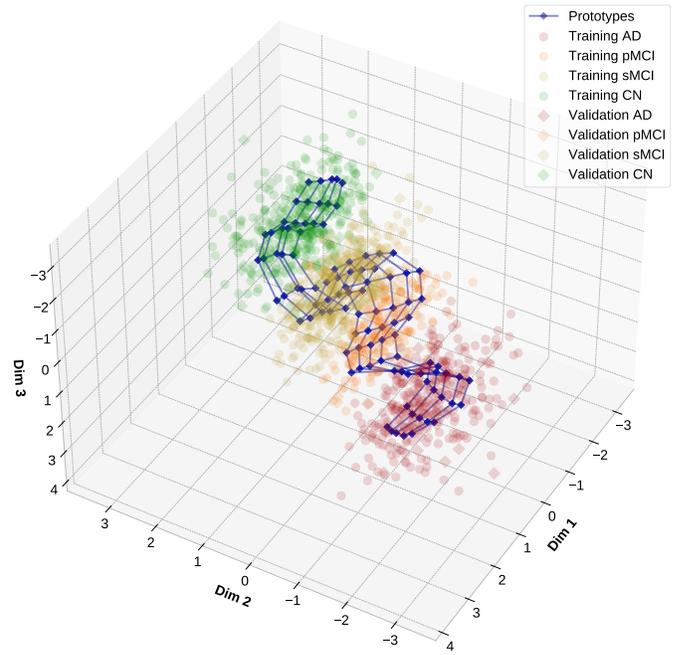

(b) 2D topology, $K = 100$

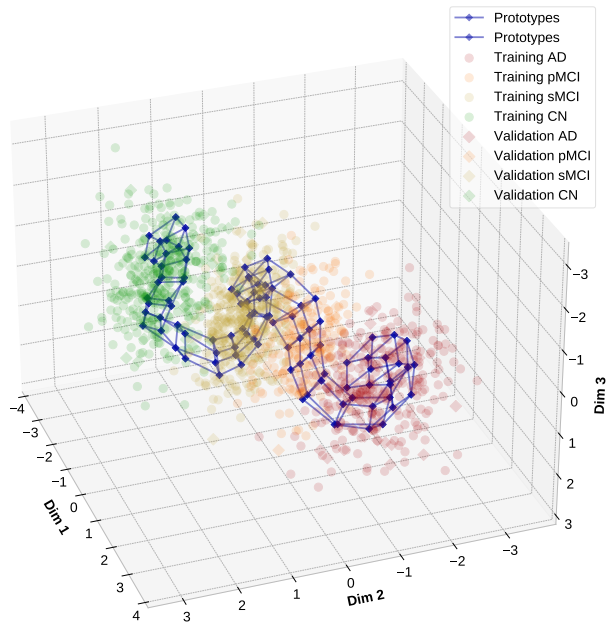

(c) 3D topology, $K = 100$

Fig. 3: Visualization of latent clinical features along with learned clinically-guided prototypes with (a) 1D, (b) 2D and (c) 3D topological arrangements using $K = 100$.

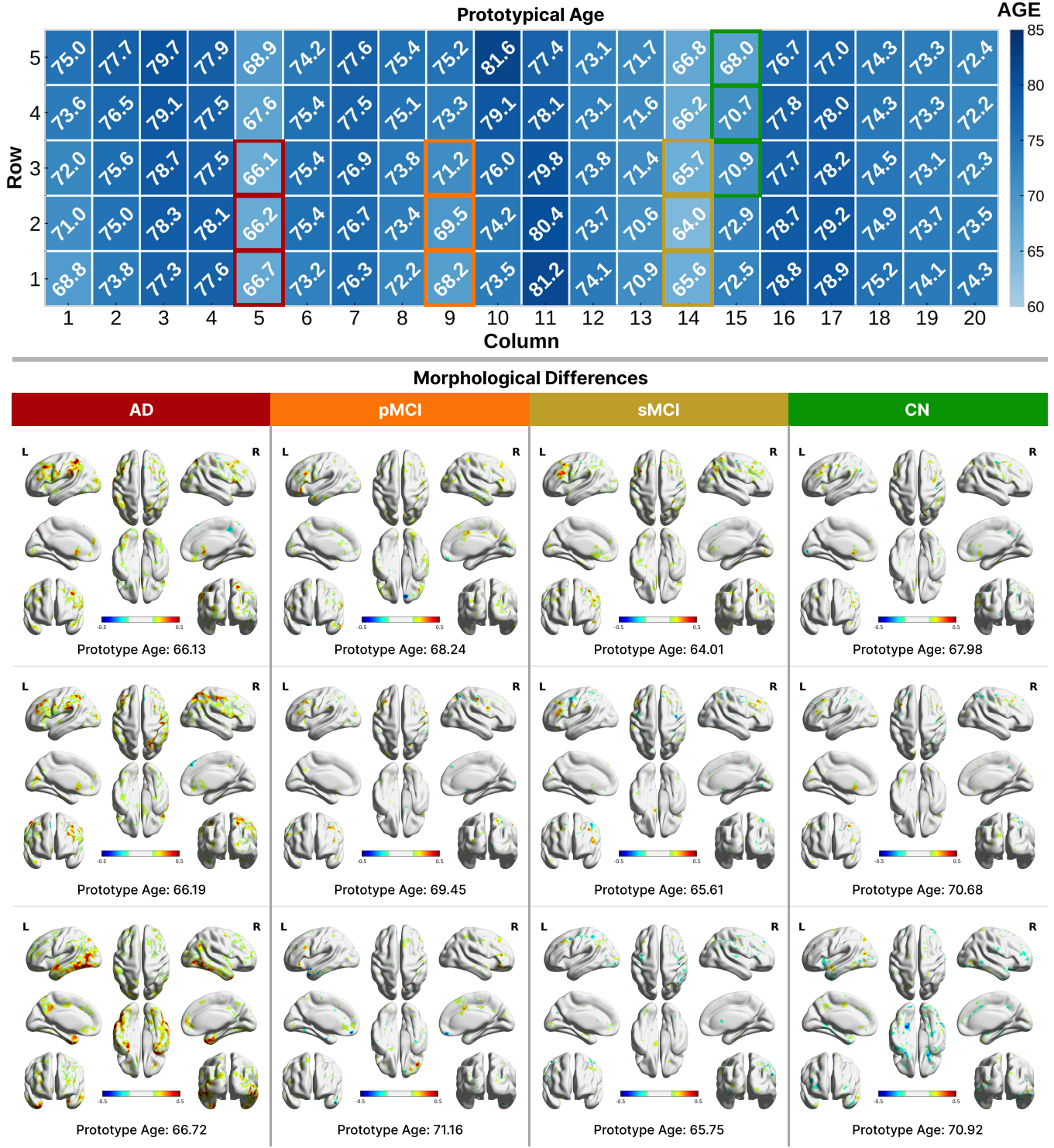

Fig. 4: Morphological changes visualization on cortical thickness by subtracting an unseen sample of an sMCI subject with an age of 61.9 to four selected prototypical brains.

**pMCI subject with an age of 55.3 (#15314)**

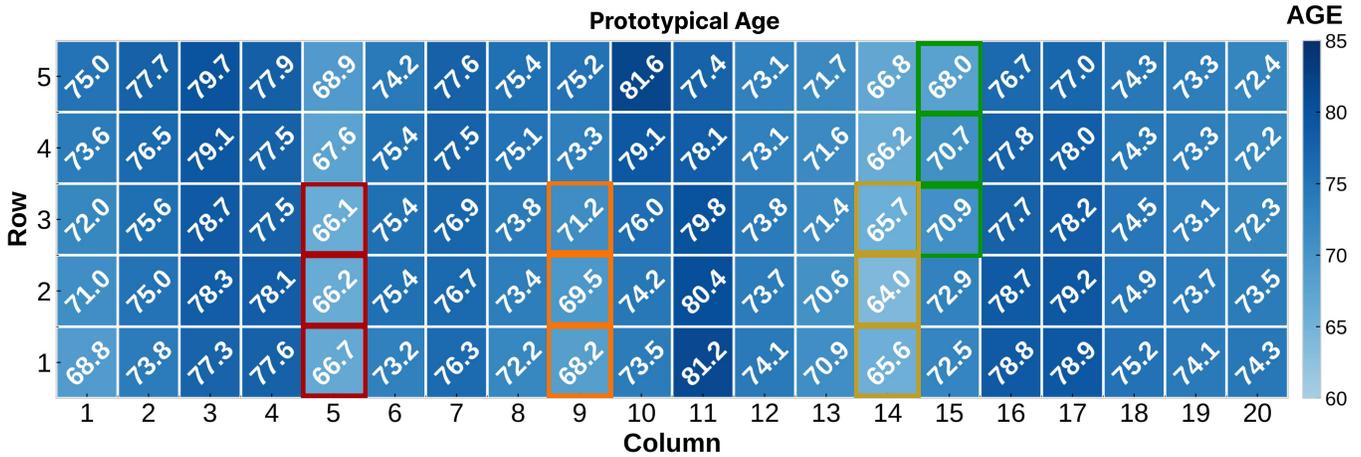

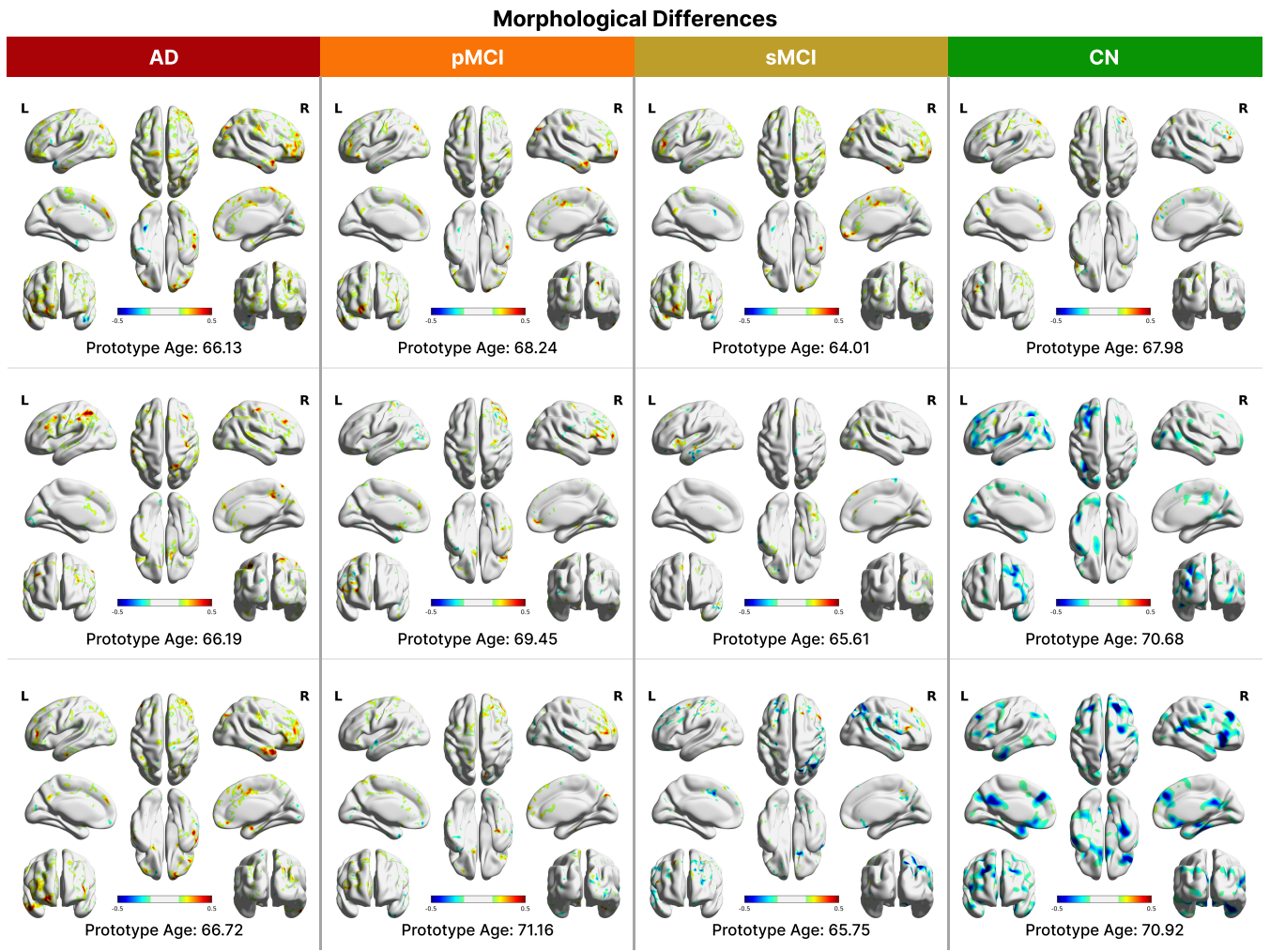

Fig. 5: Morphological changes visualization on cortical thickness by subtracting an unseen sample of a pMCI subject with an age of 55.3 to four selected prototypical brains.

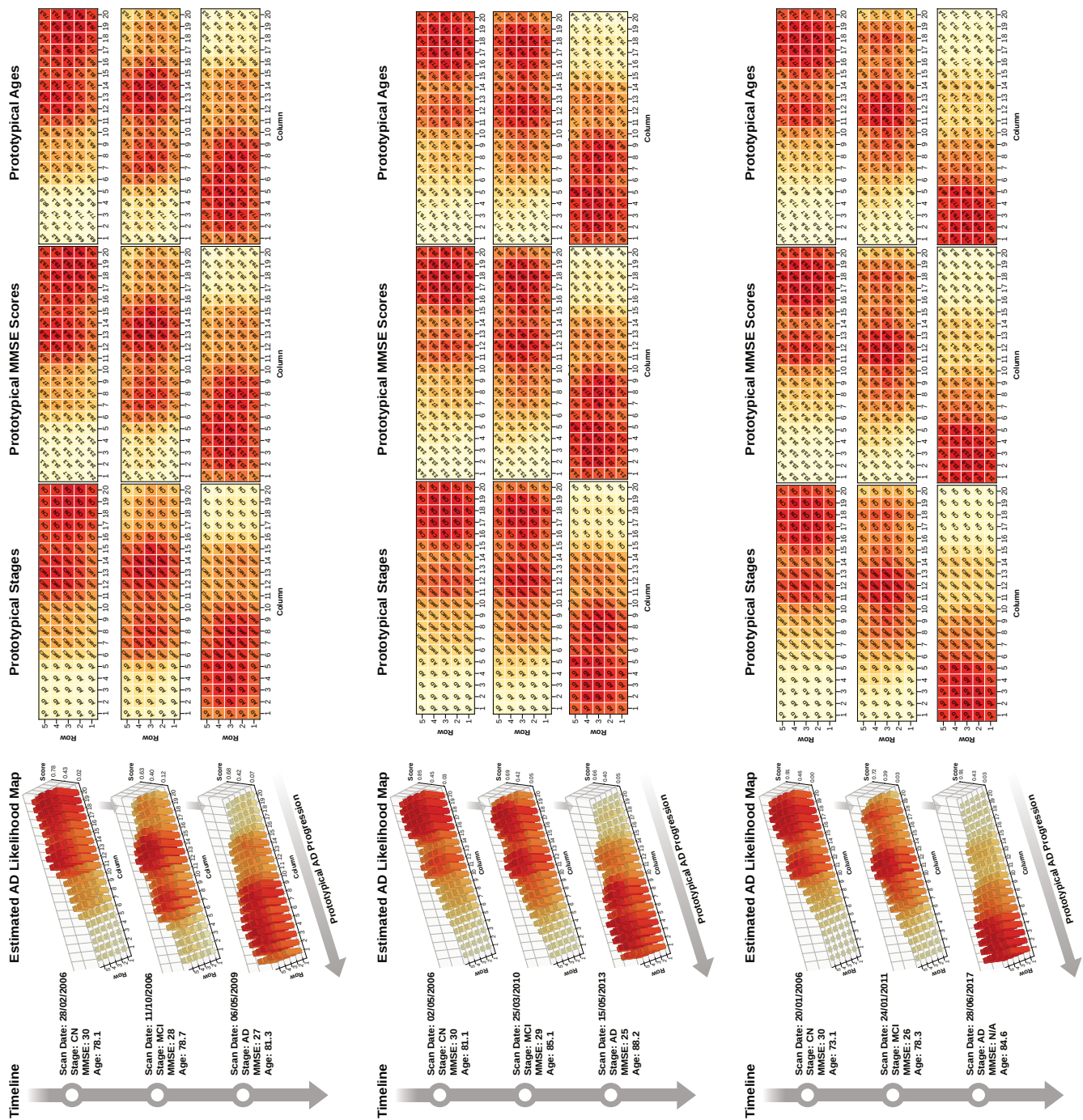

Fig. 6: Longitudinal scenarios on three healthy subjects who progressing towards AD throughout their follow-up years.

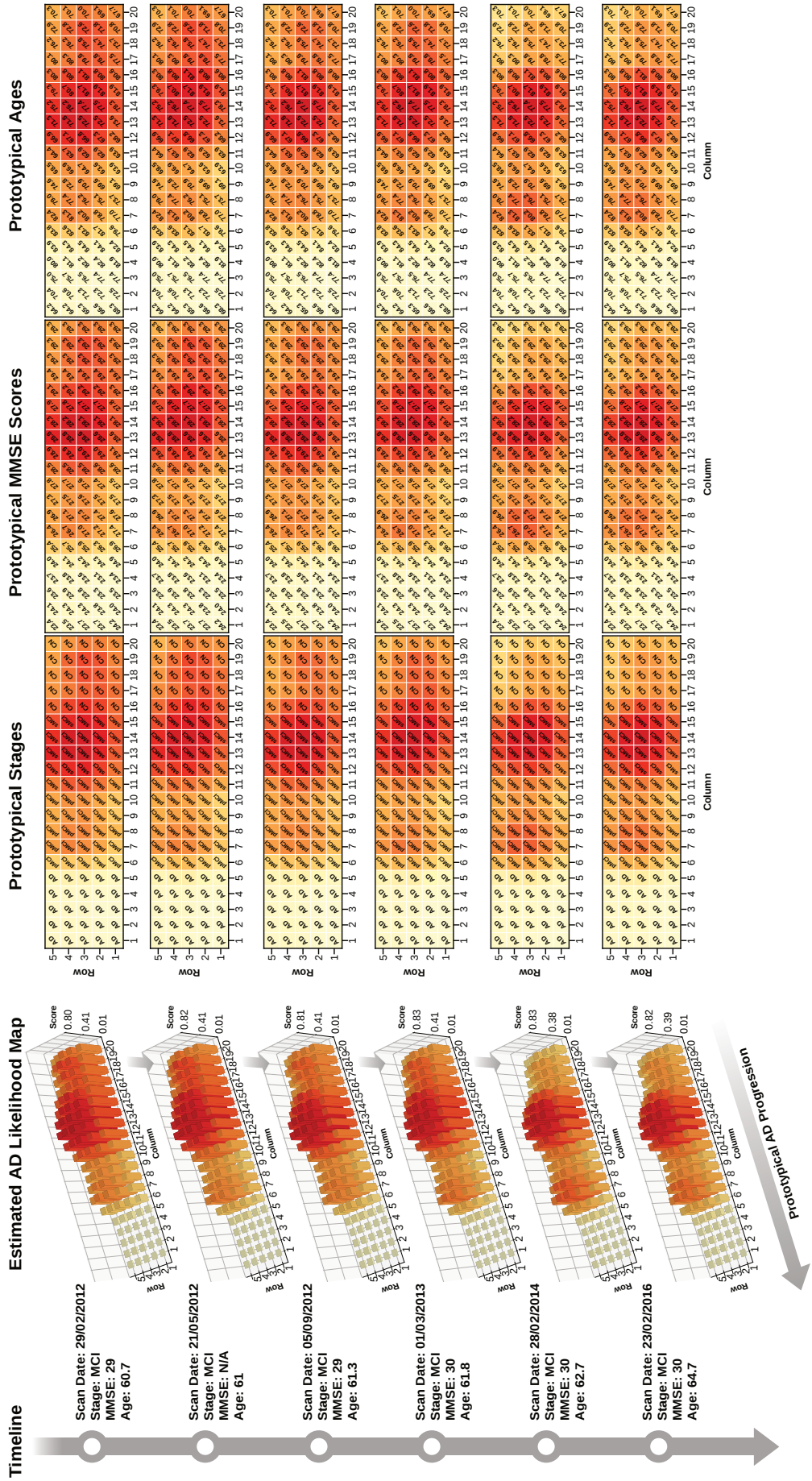

Fig. 7: Longitudinal scenario of a subject diagnosed with (stable) MCI throughout the follow-up years.

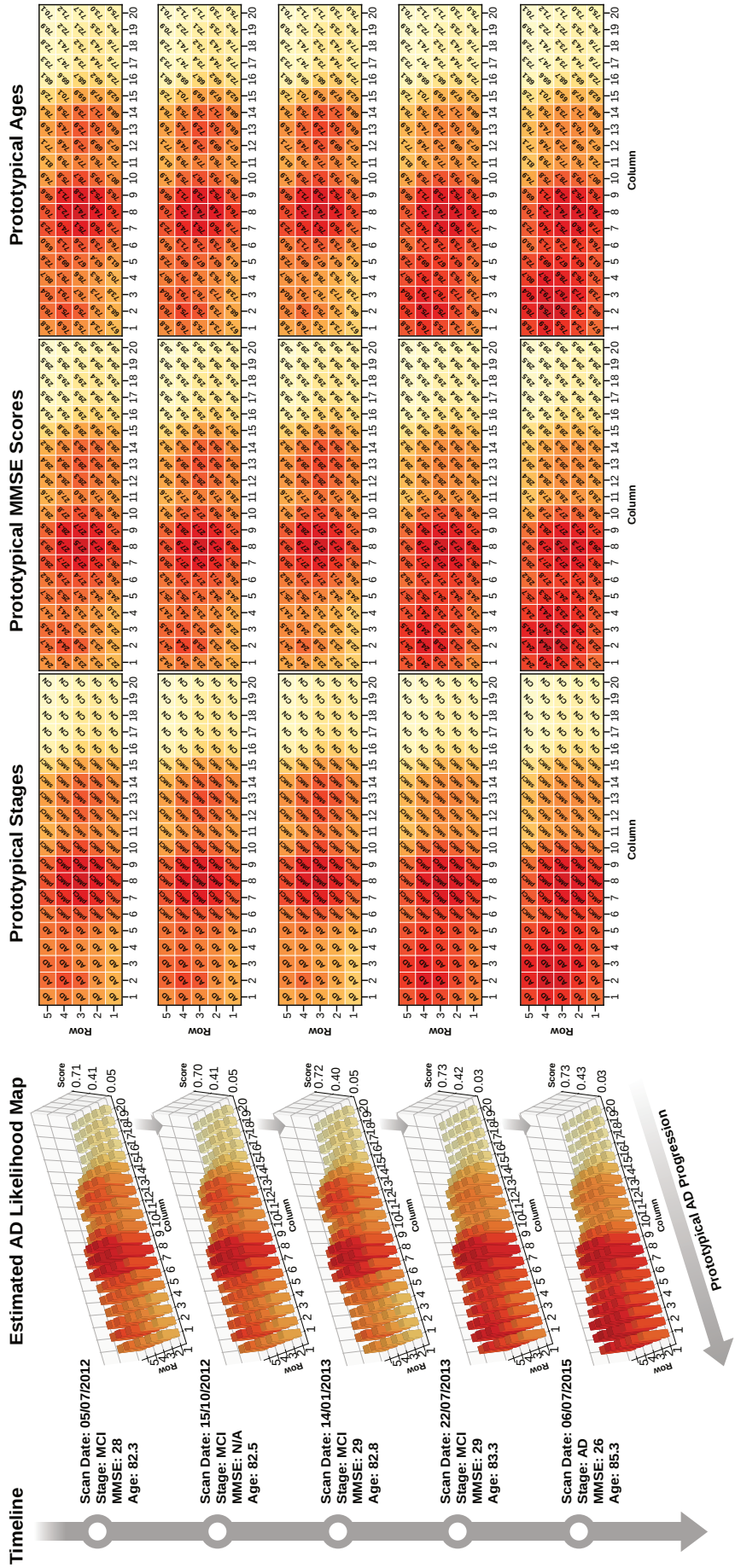

Fig. 8: Longitudinal scenario on a subject diagnosed with MCI on their first-visit scan and progressing towards AD within the following three years.



\end{document}